%% file: main_emnlp_8pg.tex
\documentclass[11pt]{article}

\usepackage{polyglossia}

\usepackage{subcaption}
\usepackage[final]{acl}

\usepackage{times}
\usepackage{tikz}
\usetikzlibrary{arrows.meta, positioning, calc, backgrounds}
\usepackage[edges]{forest}
\usepackage{graphicx}
\usepackage{times}
\usepackage{latexsym}
\usepackage{times}
\usepackage{latexsym}
\usepackage{booktabs}
\usepackage{boxedminipage}
\usepackage{amssymb}
\usepackage{amsmath}

\usepackage{graphicx}
\usepackage{enumitem}
\usepackage{listings}
\usepackage{xstring}
\usepackage{subcaption}
\usepackage{latexsym}
\usepackage{amsmath}
\usepackage{tcolorbox}
\usepackage{longtable}  %
\usepackage{booktabs}   %
\usepackage{array}      %
\usepackage{geometry}   %
\usepackage{booktabs}   %
\usepackage{tabularx}   %
\usepackage{multirow}
\usepackage{caption}
\usepackage{ragged2e}
\usepackage{tikz,xcolor,graphicx}
\usetikzlibrary{positioning,calc}
\usetikzlibrary{arrows.meta,calc}

\usepackage[T1]{fontenc}

\usepackage[utf8]{inputenc}

\usepackage{microtype}

\usepackage{inconsolata}

\usepackage{graphicx}
\usepackage{tikz}
\usepackage{forest}
\usepackage{pdflscape}  
\usepackage{booktabs}
\usepackage{tabularx}
\usepackage{array}
\usepackage{ragged2e}
\usepackage{makecell}

\usepackage{multirow}
\usepackage{array}
\usepackage{ragged2e}
\usepackage{colortbl}
\usepackage{algorithm}
\usepackage{algorithmic}

\usepackage{xcolor, rotating}
\definecolor{rowgray}{gray}{0.93}
\newcommand{\sys}[1]{\textsc{#1}}
\newcommand{\cmark}{\ding{51}}
\newcommand{\xmark}{\ding{55}}
\newcommand{\pmark}{$\bullet$}  %
\usepackage{pifont}

\newcolumntype{P}[1]{>{\RaggedRight\arraybackslash}p{#1}} %
\newcolumntype{R}[1]{>{\raggedleft\arraybackslash}p{#1}}  %

\definecolor{RowSM}{RGB}{235,245,255}      %
\definecolor{RowTPC}{RGB}{235,255,240}     %
\definecolor{RowADS}{RGB}{255,245,235}     %
\definecolor{RowTPCADS}{RGB}{245,235,255}  %

\definecolor{RowC1}{RGB}{242,248,255} %
\definecolor{RowC2}{RGB}{228,241,255}
\definecolor{RowC3}{RGB}{214,234,255}
\definecolor{RowC4}{RGB}{200,227,255} %

\usepackage{graphicx}

\usepackage{wasysym}

\definecolor{lightblue}{rgb}{.50,.95,1}
\definecolor{tri}{rgb}{.25,.88,.82}
\definecolor{lilac}{rgb}{0.85,0.64,0.85}
\definecolor{magenta}{rgb}{0.85,0.20,0.70}

\usepackage{tabularx}
\usepackage{array}

\newcolumntype{Y}{>{\raggedright\arraybackslash}X}
\newcolumntype{C}[1]{>{\centering\arraybackslash}p{#1}}

\usepackage{booktabs}
\usepackage{tabularx}
\usepackage{array}

\usepackage[table]{xcolor}
\definecolor{phasecolor}{RGB}{230,223,250}
\definecolor{totalcolor}{RGB}{226,232,240}
\definecolor{rcolor}{RGB}{230,223,200}

\lstdefinelanguage{prompt}{
  sensitive=true,
  morecomment=[l]{System},
  morecomment=[l]{ASR},
  morecomment=[l]{Dialect},
  morecomment=[l]{Emotion},
  morecomment=[l]{SSUM},
  morecomment=[l]{TSUM},
}

\usepackage{hyperref}
\usepackage{url}
\usepackage[colorinlistoftodos,prependcaption,textsize=small]{todonotes}

\presetkeys{todonotes}{inline}{}

\title{Multi-turn Conversational AI from Text to Multimodal Interaction: \\
Data, Models, Evaluation, and Open Challenges}

\author{
Syeda Faiza Ahmed Sara,
Zien Sheikh Ali,
Hunzalah Hassan Bhatti,\\
\textbf{Firoj Alam,
Shammur Absar Chowdhury} \\
Qatar Computing Research Institute, HBKU, Qatar \\
\texttt{
shchowdhury@hbku.edu.qa}
}

\setmainlanguage{english}
\setotherlanguage{arabic}

\begin{document}
\maketitle

\begin{abstract}
Conversational AI is moving beyond isolated text prompts toward sustained, multimodal interaction. In real conversations, users clarify goals, revise requests, interrupt responses, switch topics, and introduce new evidence while expecting systems to preserve context across turns. This makes multi-turn dialogue a distinct challenge requiring systems to maintain and update memory, ground responses across modalities, tools, and external knowledge, and adapt across languages and cultures. This study reviews \textbf{\textit{multi-turn conversational AI}} across text-only dialogue, AudioLLMs and speech-native systems, multimodal and omni-modal systems, and tool-augmented agents. We organize the literature around \textit{datasets} and \textit{benchmarks}, \textit{modeling paradigms}, \textit{training strategies}, \textit{evaluation setups}, and \textit{cross-cutting challenges}. Our analysis shows that support for multiple modalities has advanced faster than the ability to sustain coherent interaction across a session. Despite stronger capabilities to perceive, speak, and act across modalities, current systems still struggle with persistent memory, cross-turn grounding, full-duplex interaction, robust evaluation, and cultural alignment. We conclude with a research agenda for systems that can remember, revise, ground, speak, listen, act, and adapt across turns, modalities, and cultures.\footnote{\href{https://github.com/faiza-sfa/multiturn-conversational-ai-survey}{multiturn-conversational-resources}}
\end{abstract}

\input{sections_short/introduction}

\input{sections_short/problem_statement}

\input{sections_short/datasets}

\input{sections_short/modeling_and_challenges}

\input{sections_short/evaluation}

\input{sections_short/challenges_gaps}

\section{Conclusion}
Multi-turn conversational AI is shifting from text-only dialogue toward sustained interaction across speech, vision, video, tools, and culturally diverse contexts. This study shows that modality support has advanced faster than session-level competence. Current systems can increasingly perceive, speak, and act, but  still struggle to maintain memory, revise assumptions, preserve grounding, coordinate external state, handle spoken timing, and adapt across languages and cultures. We argue that future work should evaluate conversational systems not only by the quality of isolated responses, but by their ability to sustain coherent, grounded, safe, and culturally appropriate interaction across turns.

\section*{Limitations}
This study focuses on recent work on multi-turn conversational AI across data, models, training, and evaluation. While we aim to cover major directions, the literature is growing quickly, and some new or concurrent resources may not be included. We cover diverse systems into broad categories, which may hide finer differences in architecture, training data, deployment setting, or evaluation setups.

\section*{Societal/Broader Impact}
This study aims to clarify the capabilities, gaps, and risks of multi-turn conversational AI. By organizing prior work across datasets, models, training, and evaluation, it can help researchers design systems that better support long-horizon interaction, multilingual access, speech-based interfaces, multimodal assistance, and culturally grounded communication.

\section*{Ethical Considerations}
This study does not introduce new datasets or run experiments with human participants.

\bibliography{custom}

\appendix

\input{sections_short/appendix}

\end{document}

%% file: sections_short/introduction.tex
\section{Introduction}
\label{sec_intro}
\begin{figure}
    \centering
    \vspace{-0.3cm}
    \includegraphics[width=0.98\linewidth]{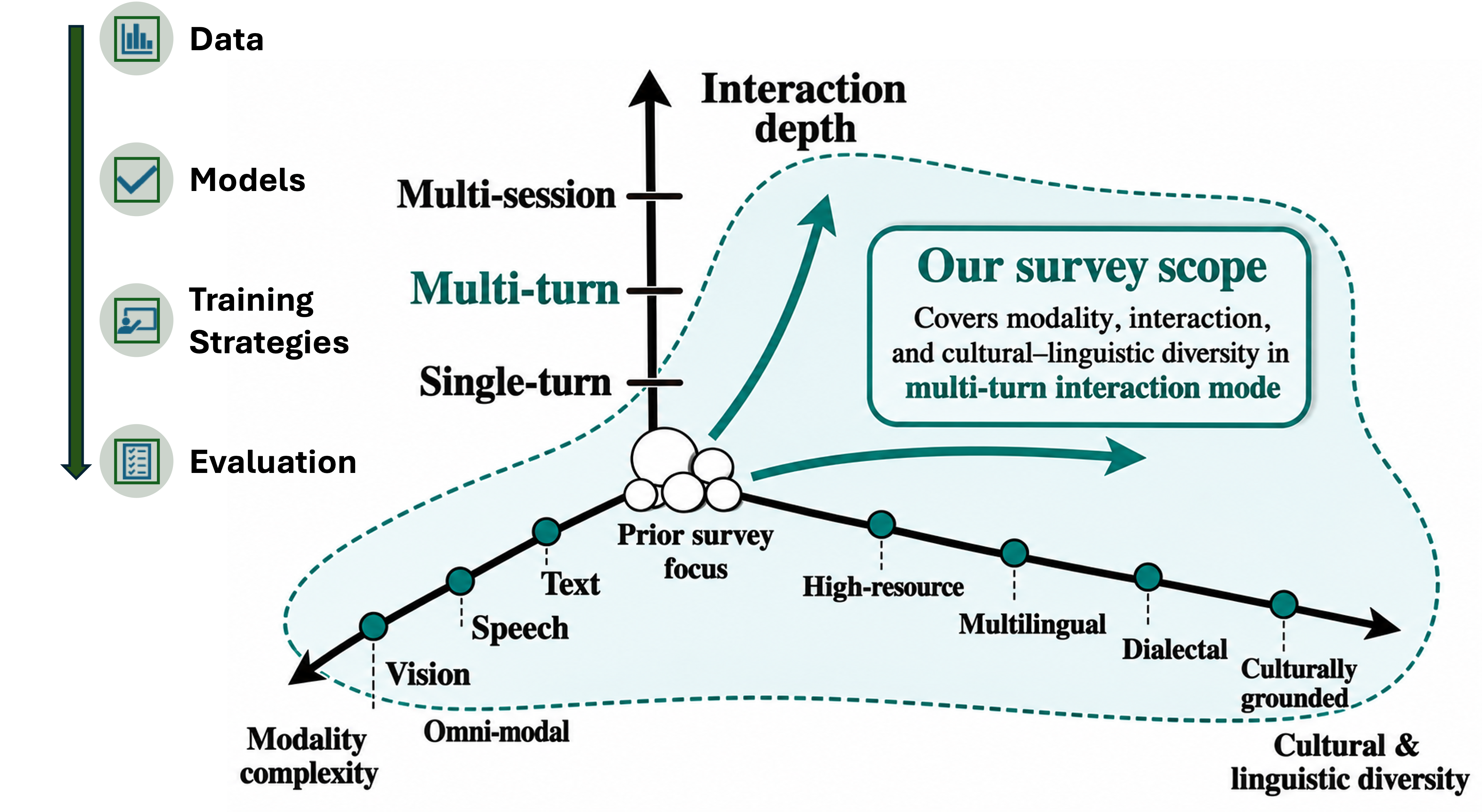}
    \vspace{-0.2cm}
    \caption{Three-axis view of conversational AI covering modality complexity, interaction depth, and cultural-linguistic diversity. 
    }
    \vspace{-0.4cm}
\label{fig:three_axes_conversational_ai}
\end{figure}
Conversational AI is becoming a primary interface through which users interact with large language models (LLMs). These interactions often extend beyond single prompts. Users clarify goals, revise requests, ask follow-up questions, switch topics, introduce new evidence, and return to earlier points in the conversation~\citep{kwan2024mteval,bai-etal-2024-mt,reddy2019coqa}. We use \textit{session-level multi-turn interaction} to refer to a complete conversation with more than one exchange, where later turns may depend on earlier requests, responses, evidence, actions, or state. A capable model must therefore do more than answer the current query. It must preserve context, resolve cross-turn references, update earlier assumptions, and maintain coherence across the session. Multi-turn dialogue is thus a distinct modeling and evaluation problem, not merely a longer version of single-turn question answering~\citep{deshpande-etal-2025-multichallenge}.

This problem has become more important as conversational AI expands beyond text. Early dialogue systems focused on text-based state tracking, response generation, persona consistency, and task completion~\citep{wu2020tod,zhang-etal-2020-dialogpt,roller-etal-2021-recipes}. AudioLLMs extend this setting to spoken interaction by processing speech with text~\citep{zhang2023speechgpt,chu2024qwen2,tang2024salmonn}. Full-duplex systems support more flexible turn-taking, including real-time interruption and overlapping speech~\citep{defossez2024moshi}. More recent omni-modal models jointly handle text, speech, vision, and often video within unified architectures~\citep{hurst2024gpt,qwen25omni,fu2026vita,chen2025emova,tong2025interactiveomni}. Figure~\ref{fig:evolution_complexity} illustrates this progression. Each stage broadens what 
systems can perceive and produce, while %
making memory, grounding, timing, and alignment across turns more difficult.

Recent evidence shows that even frontier models still struggle with sustained interaction. Models can underuse relevant information even when it remains inside the context window~\citep{liu-etal-2024-lost}. They also degrade when a task is distributed across several turns rather than stated as a complete single-turn instruction~\citep{laban2025lostmultiturn}. MultiChallenge further shows that frontier models remain below human-level reliability on realistic multi-turn instruction following~\citep{deshpande-etal-2025-multichallenge}. These failures become more complex in spoken, multimodal, and tool-augmented settings. Speech recognition or acoustic-token errors can propagate across turns, visual grounding can decay as the dialogue moves away from the original image or video, and tool-using agents must preserve external state in addition to dialogue history. Evaluation setups that score turn-level responses can therefore miss failures that only emerge at the session level~\citep{kwan2024mteval,graf2026turnwise}.

\paragraph{Related Surveys.} 
Prior surveys provide useful coverage of dialogue systems, multi-turn LLM capabilities, agents, and multimodal LLMs, however, they study these areas largely in isolation. Dialogue-system surveys review the transition from modular or retrieval-based systems to LLM-based dialogue, however, remain largely text-centered~\citep{wang2023survey,yi2025survey}. Surveys on multi-turn interaction discuss instruction following, memory, and consistency, however, do not cover spoken or multimodal interaction in depth~\citep{zhang2025survey,li2025beyond}. Work on LLM agents focuses mainly on tool use and agent evaluation~\citep{guan2026evaluating}, while surveys of multimodal LLMs often treat dialogue as a downstream application rather than the central unit of analysis~\citep{zhang2024mm}. Cultural and dialectal variation is also largely absent from these discussions. In contrast, this survey treats \textbf{sustained interaction} as the \textit{primary unit of analysis} across modalities, systems, resources, and evaluation methods. Table~\ref{tab:survey-positioning} summarizes how our scope differs from the closest related surveys. Appendix Table~\ref{tab:survey-positioning-detailed} provides a more detailed comparison of their conceptual framing, modality coverage, and key distinctions.

\begin{table}[t]
\centering
\small
\setlength{\tabcolsep}{3pt}
\scalebox{0.80}{%
\begin{tabular}{l c c c c c c}
\toprule
\textbf{Survey} & \textbf{MT} & \textbf{Speech} & \textbf{Vision} & \textbf{Tri} & \textbf{Cultural} & \textbf{Eval$\Delta$} \\
\midrule
\sys{\cite{yi2025survey}}
  & \cmark & \xmark & \xmark & \xmark & \xmark & \cmark \\
\sys{\cite{wang2023survey}}
  & \cmark & \xmark & \xmark & \xmark & \xmark & \xmark \\
\sys{\cite{zhang2025survey}}
  & \cmark & \xmark & \xmark & \xmark & \xmark & \cmark \\
\sys{\cite{li2025beyond}}
  & \cmark & \xmark & \xmark & \xmark & \pmark & \cmark \\
\sys{\cite{guan2026evaluating}}
  & \cmark & \xmark & \xmark & \xmark & \xmark & \cmark \\
\sys{\cite{zhang2024mm}}
  & \xmark & \pmark & \cmark & \xmark & \xmark & \xmark \\
\midrule
\textbf{This survey}
  & \cmark & \cmark & \cmark & \cmark & \cmark & \cmark \\
\bottomrule
\end{tabular}
}
\vspace{-0.2cm}
\caption{%
Comparison with closely related surveys. \textbf{MT} = multi-turn focus; \textbf{Speech} = spoken dialogue/AudioLLMs; \textbf{Vision} = image/video; \textbf{Tri} = joint text--speech--vision; \textbf{Cultural} = cross-lingual, dialectal, or culturally grounded coverage; \textbf{Eval$\Delta$} = multi-turn evaluation gaps. \pmark{} denotes partial coverage.%
}

\label{tab:survey-positioning}
\vspace{-0.4cm}
\end{table}

\paragraph{Survey Scope and Selection.}
We organize our coverage along three dimensions: \textit{interaction depth} (multi-turn sessions where later responses depend on earlier context), \textit{modality complexity} (text through speech to omni-modal systems), and \textit{cultural and linguistic diversity} (multilingual, dialectal, and culturally grounded settings). Figure~\ref{fig:three_axes_conversational_ai} illustrates this joint space. We include work that advances dialogue across turns or develops the \textbf{data}, \textbf{models}, \textbf{training methods}, and \textbf{evaluation} needed for sustained interaction, spanning topics such as \textit{text dialogue, spoken dialogue, AudioLLMs, vision-language and video dialogue, omni-modal systems, conversational retrieval, tool use, and culturally grounded resources}. Details keywords are provided in Appendix Sec. \ref{app:keywords}.
To ensure a systematic, and reproducible review of the literature, we followed \textit{PRISMA-ScR} framework~\citep{page2021prisma}. To maximize coverage, we searched Google Scholar and Semantic Scholar and obtained papers from IEEE Xplore, ACM Digital Library, Elsevier, DBLP, the ACL Anthology and arXiv. Papers include publications from $^*$ACL, NeurIPS, ICLR, ICML, Interspeech, SIGDIAL, CVPR, ICCV, AAAI and related venues. Our initial search identified $\sim$4K papers. After applying the PRISMA screening, we selected 200 papers for detailed review.

\noindent\textbf{Contributions.} This survey makes following contributions.
\begin{itemize} [noitemsep,topsep=0pt,leftmargin=1.5em,labelsep=.5em]
    \item We provide a unified survey of multi-turn conversational AI spanning
text, speech, vision, video, agentic, and culturally grounding.  %

    \item We organize datasets, benchmarks, models, training strategies, and evaluation methods around session-level interaction rather than isolated responses.

    \item We identify open challenges in memory, coherence, paralinguistic understanding, multimodal grounding, real-time interaction, and cultural alignment.
\end{itemize}

\paragraph{Findings.} 
This survey highlights several gaps in the current multi-turn conversational AI.
\begin{itemize}[noitemsep,topsep=0pt,leftmargin=1.5em,labelsep=.5em]
    \item \textbf{Capability gap.} Longer context windows, stronger base models, and broader modality support do not guarantee session-level competence. Systems still struggle with memory, grounding, assumption revision, tool state, spoken timing, and cross-cultural adaptation.

    \item \textbf{Resource gap.} Existing datasets remain dominated by English with (over 80\%), text-only, and text-image interaction. Speech, video, omni-modal, %
    and culturally grounded multi-turn resources remain limited.

    \item \textbf{Evaluation gap.} Benchmarks increasingly use LLM-as-judge and mixed approaches, but session-level metrics, human validation, and reproducible scoring remain underdeveloped.

    \item \textbf{Integration gap.} Most resources test one dimension at a time. Few combine long-horizon memory, multilinguality, speech, visual grounding, tool use, safety, and cultural alignment in the same evaluation setting.
\end{itemize}

%% file: sections_short/problem_statement.tex
\section{Multi-turn Dialogue} %

\label{sec:problem_definition_scope}

We define a \textit{\textbf{conversation} as an interaction episode between a user and a system, consisting of ordered turns}. A \textit{\textbf{multi-turn conversation} is a session with \(T>1\) exchanges, where later turns may depend on earlier requests, responses, evidence, actions, or state}. This dependency makes the full session the natural unit of analysis rather than a single response.
\noindent We define a session, $\mathcal{D}$, as

\[
\small
\begin{aligned}
\mathcal{D} &= \{(u_t,y_t,c_t,m_t,z_t)\}_{t=1}^{T},\\[-2pt]
y_t &= f_\theta(u_t,c_t,m_t,z_t).
\end{aligned}
\]
Here \(u_t\) is the user input at turn \(t\), \(y_t\) is the system response, \(c_t\) is the accumulated dialogue context from earlier turns, \(m_t\) is the multimodal context, and \(z_t\) is optional external state. The multimodal context may include speech, images, video, or other non-text signals, and may be empty in text-only settings. The external state may include memory, retrieved evidence, tool outputs, task state, or user profile information. The function \(f_\theta\) denotes the conversational model. This formulation makes the multi-turn setting explicit because each response can depend on the current input, the previous dialogue, modality-specific context, and external state.

\paragraph{Modality and Context.}
The user input \(u_t\) and system response \(y_t\) may be expressed as \textit{text}, \textit{speech}, \textit{images}, \textit{video}, or \textit{their combination}. The multimodal context \(m_t\) captures modality specific information such as acoustic cues, visual content, or video events. The dialogue context \(c_t\) captures what has already happened in the session, while the external state \(z_t\) captures information that may persist or change. Since these signals carry different information, the model must decide what to retain, update, or ignore at each turn.

\paragraph{Dialogue Types.}
Dialogue systems serve different interaction functions. We consider \textbf{task oriented dialogue} for goal completion via state tracking~\cite{wu2020tod,eric2020multiwoz}, \textbf{open-domain dialogue} for coherent conversation across topics~\cite{zhang-etal-2020-dialogpt,roller-etal-2021-recipes}, \textbf{knowledge grounded dialogue} for responses grounded in documents, databases, or retrieval~\cite{varshney2022commonsense}, \textbf{social and empathetic dialogue} for socially appropriate responses~\cite{zhang2024towards}, \textbf{persona grounded dialogue} for consistency with a user or character profile~\cite{gosling2023pippa,oh-etal-2023-pk}, \textbf{agentic dialogue} for planning and tool use~\cite{yao2022react,pmlr-v235-lu24e}, and \textbf{multimodal grounded dialogue} for reasoning over text, speech, images, video, or their combination~\cite{xue2025mmrc}. These types often overlap, so they serve as analytical lenses  %
not strict dataset boundaries.

\paragraph{Evaluation Focus.}
A multi-turn system is mainly evaluated at the \textbf{turn} and \textbf{session} levels. Turn level evaluation asks whether the current response is correct and useful. Session level evaluation asks whether the system remains consistent, grounded, and useful as earlier turns shape later ones. Some settings add further requirements. Persistent memory extends the problem beyond one session, tool using systems add external actions and state changes, and spoken systems may require full duplex evaluation for interruptions, overlapping speech, and response timing. Our focus is interaction within a multi-turn session, while cross session memory and agentic interaction are covered as related extensions when they affect session level behavior.

%% file: sections_short/datasets.tex
\section{Datasets and Benchmarks}
\label{sec:dataset}

Datasets and benchmarks for multi-turn dialogue differ not only in modality, but also in what they assume a system must preserve across turns. Text resources emphasize context tracking, instruction retention, persona consistency, retrieval, and social intelligence. Spoken resources add acoustic cues, turn-taking, prosody, speaker variation, ASR errors, and full-duplex timing. Multimodal and video resources add visual, audio-visual, and temporal grounding. Cultural and linguistic resources expose whether these abilities transfer beyond English-centered settings.

\input{tables/dataset_table}

\input{tables/compact_benchmark_table}

We organize datasets by modality in Table~\ref{tab:datasets-mt} and benchmarks by evaluation focus in Table~\ref{tab:benchmarks-mt}. Overall, they cover text-only, spoken, multimodal, video, and culturally grounded interaction. Some findings are as follows.

\begin{itemize}[noitemsep,topsep=0pt,leftmargin=1.5em,labelsep=.5em]
    \item \textbf{English and text remain dominant.}
    Outside the cultural and cross-lingual block, 43 of 52 dataset entries are English-only. The benchmark table shows the same pattern, with 32 of 53 benchmarks focusing on text-only dataset.
    \item \textbf{Modality coverage is expanding, however, varies across modalities.} %
    Table~\ref{tab:datasets-mt} lists 18 text-image resources, however, only 8 spoken and 6 video-oriented datasets. Many multimodal resources still use static images rather than streaming, temporal, or spoken interaction.
    \item \textbf{Task coverage is broadening.}
    Though QA and instruction following dominate, newer resources add memory, personalization, tool use, retrieval, social interaction, and emotional intelligence. Long-horizon and multi-session dialogue remain  rare.

    \item \textbf{Cultural coverage is growing, however, often single-turn.}
    Resources such as CVQA, Dallah, MMA-ASIA, OASIS, and Shawarma Chats improve cultural coverage. However, 4 of the 5 entries in this block are single-turn and do not test sustained interaction.

    \item \textbf{Evaluation needs stronger calibration.}
    LLM-as-judge and mixed scoring now dominate many benchmarks. These setups scale well, but require stronger human evaluation and clearer reproducibility. Safety and robustness evaluation also remains text-centered.
\end{itemize}

%% file: tables/dataset_table.tex
\begin{table*}[!ht]
\centering
\small
\setlength{\tabcolsep}{2.5pt}
\scalebox{0.70}{
\begin{tabular}{l c l c r l l l}
\toprule
\textbf{Reference}
  & \textbf{Mod.}
  & \textbf{Task}
  & \textbf{Turns}
  & \textbf{Size}
  & \textbf{Lang.}
  & \textbf{Split} %
  & \textbf{Curation} \\
\midrule

\multicolumn{8}{c}{\textbf{Text-only}} \\
\midrule

\sys{PersonaChat}~\cite{zhang2018personachat}
  & T & PS   & $6$--$20$ & 10{,}981/164K$^{u}$     & en      & Tr    & H   \\
\sys{CoQA}~\cite{reddy2019coqa}
  & T & QA   & $6$--$20$ & 8{,}399        & en      & Tr+Ev    & H   \\
\sys{MultiWOZ2.1}~\cite{eric2020multiwoz}
  & T & TOD/DST & $6$--$20$. & 10{,}438 & en & Tr+Ev & H \\
\sys{PIPPA}~\cite{gosling2023pippa}
  & T & PS   &  $>20$        & 25{,}940/$>$1M$^{u}$    & en      & Tr    & H+S \\
\sys{UltraChat}~\cite{ding2023ultrachat}
  & T & Inst &    ${\leq}5$     & 1.5M           & en      & Tr    & S   \\
\sys{MT-Bench}~\cite{zheng2023judging}
  & T & Inst & ${\leq}5$ & 80$^{q}$           & en      & Ev    & H   \\
\sys{WildChat}~\cite{zhao2024wildchat}
  & T & OD   & ${\leq}5$ & 1M             & ML(68)      & Tr+Ev & H   \\
\sys{MT-Bench-101}~\cite{bai-etal-2024-mt}
  & T & Inst & ${\leq}5$ & 1{,}388        & en      & Ev    & H+S \\
\sys{MT-Eval}~\cite{kwan2024mteval}
  & T & Inst & $6$--$20$ & 168            & en      & Ev    & H+S   \\
\sys{PRODIGy}~\cite{occhipinti2024prodigy}
  & T & PS   & 4      &  20{,}850             & en      & Tr+Ev    & H+S   \\
\sys{LongMemEval}~\cite{wu2025longmemeval}
  & T & Mem  & MS        & 500$^{q}$      & en      & Ev    & H+S   \\
\sys{MultiChallenge}~\cite{deshpande-etal-2025-multichallenge}
  & T & Inst & $\leq$10 / 5 avg & 273        & en      & Ev    & H+S \\
\sys{LMSYS-Chat-1M}~\cite{zheng2024lmsyschatm}
   & T & OD & $\leq$5 & 1M & ML(154) & Tr+Ev & H+S \\
\sys{$\tau$-Bench}~\cite{yao2025tau}
   & T & TOD & -- & 165tasks & en & Ev & S+H \\
\sys{PersonaMem}~\cite{jiang2025know}
  & T & Pers/Mem & MS
  & 180+ histories / $\sim$6K$^{q}$ & en & Ev & S+H \\
\sys{ConsistentChat}~\cite{chen2025consistentchat} & T & Inst & 6--20 & 15K / 224K$^{u}$ & en & Tr & S \\
\sys{DocTalk}~\cite{lee-etal-2025-doctalk} & T & KG/QA & ${>}20$ & 730K & en & Tr & H+S \\
\sys{ToolWOZ}~\cite{lattimer-etal-2025-sparse} & T & TOD+Tool & -- & 7,849 scenarios & en & Tr+Ev & H+S \\
\sys{SOTOPIA}~\cite{zhou2024sotopia} & T & Social/PS & 6--20 & 450 tasks & en & Ev & H+S \\
\sys{DialSim}~\cite{kim2024dialsim} & T & Mem & $>20$ & 18.99K turns / 1.31K sess. & en & Ev & S+H \\

\midrule
\multicolumn{8}{c}{\textbf{Spoken (audio\,/\,speech)}} \\
\midrule

\sys{SpokenWOZ}~\cite{si2023spokenwoz}
  & T+S & TOD/DST  & $>20$      & 5.7K\,/\,249$^{h}$ & en & Tr+Ev & H \\
\sys{DeepDialogue}~\cite{deepdialogue2025}
  & T+S & OD/EI   & $3$--$10$ & 40{,}150     & en      & Tr    & S + H   \\
\sys{Audio MultiChallenge}~\cite{gosai2025audiomultichallenge}
  & S   & Inst & $6$--$20$ & 452          & en      & Ev    & H+S \\
\sys{MENASpeechBank}~\cite{ali2026menaspeechbankreferencevoicebank}
  & T+S   & TOD/Pers   & ${\leq}5$  & 417K    & ar*     & Tr+Ev & H+S   \\
\sys{C3}~\cite{ma-etal-2025-c3}
  & T+S & QA & 6--20 & 1,079 & en+zh & Ev & H+S \\
\sys{ASK-QA}~\cite{chen-etal-2025-data} & T+S & QA & ${\leq}$5 & 7{,}830 & en & Tr+Ev & S \\
\sys{Multi-Bench}~\cite{deng2025multibench} & S & OD+EI & 6--20 & 1.5K & en+zh & Ev & H+S \\
\sys{MSIB}~\cite{tong2025interactiveomni} & T+S & OD & $6$--$20$ & 244 & en & Ev & S+H \\
\midrule
\multicolumn{8}{c}{\textbf{Multimodal}} \\
\midrule

\sys{VisDial}~\cite{das2017visual}
  & T+V & QA   & $6$--$20$ & $\sim$123K / 1.23M$^{q}$        & en      & Tr+Ev & H   \\
\sys{MMDialog}~\cite{feng2023mmdialog}
  & T+V & OD   & ${\leq}5$      & 1.08M          & en      & Tr    & H   \\
\sys{IMAD}~\cite{moskvoretskii2024imad}
  & T+V & OD   & $6$--$20$ & 4864         & en      & Tr    & H   \\
\sys{InfoVisDial}~\cite{wen2023infovisdial}
  & T+V & KG   & $6$--$20$ & 50K          & en      & Tr    & S   \\
\sys{DialogCC}~\cite{lee2024dialogcc} 
    & T+V & OD & $6$--$20$ & 83K & en & Tr & H+S \\
\sys{LoCoMo}~\cite{maharana2024evaluating}
  & T+V & Mem & ${>}20$  & 10 / 1{,}986$^{q}$ & en      & Ev    & H+S \\
\sys{TMDialog}~\cite{lei2025contextqformer} & T+V & Inst & $6$--$20$ & 67.9K / 329 & en & Tr+Ev & H+S \\
\sys{MMDU-45k}~\cite{liu2024mmdu}
  & T+V & Inst & $6$--$20$ & 45K / 410K$^{q}$    & en      & Tr    & S+H \\
\sys{ConvBench}~\cite{liu2024convbench}
  & T+V & Inst & ${\leq}5$  & 577          & en      & Ev    & H+S \\
\sys{CB-300K}~\cite{tian2025chatterbox}
  & T+V & QA   & ${\leq}5$ & 340k/717K$^{q}$  & en      & Tr+Ev    & H+S \\
\sys{MMDiag}~\cite{liu2025diagnote}
  & T+V & QA & ${\leq}5$ & 639K$^{q}$          & en      & Tr+Ev & H+S \\
\sys{MultiVerse}~\cite{lee2025multiverse}
  & T+V & Inst & ${\leq}5$ & 647          & en      & Ev    & H+S \\
\sys{MMRC}~\cite{xue2025mmrc}
  & T+V & Mem+QA & $6$--$20$ & 5{,}120\,/\,28K$^{q}$ & en & Ev & H \\
\sys{AlignMMBench}~\cite{wu2025alignmmbench}
  & T+V & Inst & --      & 4{,}978$^{q}$             & zh      & Ev    & H+S \\
\sys{Mem-Gallery}~\cite{bei2026mem}
  & T+V & Mem  & MS      & 240 sess. / 1{,}711$^{q}$             & en      & Ev    & H+S \\
\sys{MMMB}~\cite{tong2025interactiveomni} & T+V & Mem & $6$--$20$ & 300 & en & Ev & S+H \\

\sys{DialogBen}~\cite{huang-etal-2025-dialoggen} 
  & T+V & T2I & ${\leq}5$ & 9,957 & en+zh 
  & Ev & S \\
\sys{MMMT-IF}~\cite{epstein2024mmmtif} & T+V & Inst & 1--20 & 71 & en & Ev & H+S \\

\midrule
\multicolumn{8}{c}{\textbf{Video (text\,+\,video\,/\,streaming)}} \\
\midrule

\sys{MT-Video-Bench}~\cite{pan2025mtvidobench}
  & T+Vid   & QA   & $6$--$20$ & 1{,}000 / 5{,}887$^{q}$       & en      & Ev    & H+S \\
\sys{OmniMMI}~\cite{wang2025omnimmi}
  & T+S+Vid & QA & ${\leq}5$ & 1{,}121$^{v}$/2{,}290$^{q}$      & en      & Ev    & H+S \\
\sys{SCVBench}~\cite{ijcai2025p255}
  & T+Vid   & QA   & $6$--$20$ & 925$^{v}$ / 7{,}280$^{q}$ & en    & Ev    & H+S \\
\sys{CogStream}~\cite{zhao2026cogstream}
  & T+Vid   & QA   & --- & 1{,}088$^{v}$ / 59{,}032$^{q}$      & en      & Tr+Ev    & H+S \\
\sys{IVCR-200K}~\cite{han2024ivcrk} & T+Vid & Ret & 6--20 & 12{,}516$^{v}$ / 201{,}631$^{q}$ & en+zh & Tr+Ev & S+H \\
\sys{SVBench}~\cite{yang2025svbench} & T+Vid & QA & ${\leq}5$ & 49,979$^{q}$ / 1,353$^{v}$ & zh & Tr+Ev & H+S \\

\midrule
\multicolumn{8}{c}{\textbf{Cultural\quad($^{\star}$\,single-turn)}} \\
\midrule

\sys{CVQA}$^{\star}$~\cite{mogrovejo2024cvqa}
  & T+V     & QA & 1  & 10{,}374$^{q}$      & ML(31)  & Ev    & H   \\
\sys{Dallah}$^{\star}$~\cite{alwajih2024dallah}
  & T+V     & QA & 1 & 20$\times$6$^{q}$        & ar(6D)     & Ev    & H   \\
\sys{MMA-ASIA}$^{\star}$~\cite{weihua2025mmaasia}
  & T+S+V   & QA & 1 & 27{,}000$^{q}$        & ML(10)  & Ev    & H+S \\
\sys{OASIS}$^{\star}$~\cite{alam2025everydaymmqa}
  & T+S+V   & QA & $1$ & 14.8M$^{q}$ & en+ar* & Tr+Ev  & H+S \\
\sys{Shawarma Chats}~\cite{zeinalipour-etal-2025-shawarma} & T & OD+KG & 6--20 & 30K & ar*(3D) & Tr+Ev & H+S \\

\bottomrule
\end{tabular}
}
\vspace{-0.3cm}
\caption{%
  Multi-turn dialogue \textbf{datasets} %
  organised by modality.
  \textbf{Mod.}:
    T\,=\,text,
    S\,=\,speech/audio,
    V\,=\,image/vision,
    Vid\,=\,video.
  \textbf{Task}:
    TOD\,=\,task-oriented dialogue,
    OD\,=\,open-domain,
    QA\,=\,question answering,
    PS\,=\,persona / roleplay,
    Pers.\,=\,personalization / user profiling,
    KG\,=\,knowledge-grounded,
    Inst\,=\,instruction-following benchmark,
    T2I\,=\,text-to-image generation / editing,
    EI\,=\,emotional intelligence,
    Ret \,=\, retrieval,
    Mem\,=\,long-term memory.
  \textbf{Turns}: approximate range per session
    (${\leq}5$ / $6$-$20$ / ${>}20$ / MS\,=\,multi-session).
  \textbf{Size}: \#\,dialogues unless marked:
    $^{u}$\,=\,utterances;
    $^{q}$\,=\,QA pairs\,/\,instances;
    $^{h}$\,=\,hours;
    $^{v}$\,=\,videos.
  \textbf{Split}:
    Tr\,=\,training data,
    Ev\,=\,evaluation data,
    Tr+Ev\,=\,both.
  \textbf{Curation}:
    H\,=\,human;
    S\,=\,synthetic\,/\,LLM-generated;
    \textbf{Lang.}:
    ML(*)\,=\,Multilingual; Number in parenthesis indicate number of languages.
    D\,=\,Dialects.
  Datasets marked $^{\star}$ are \emph{single-turn};
  they are included solely as cultural\,/\,cross-lingual %
  baselines.
}
\label{tab:datasets-mt}
\vspace{-0.35cm}
\end{table*}

%% file: tables/compact_benchmark_table.tex
\begin{table}[!ht]
\centering
\setlength{\tabcolsep}{2.0pt}
\scalebox{0.58}{
\begin{tabular}{l l c l}
\toprule
\textbf{Benchmark} & \textbf{Mod.} & \textbf{Turns} & \textbf{Eval} \\
\midrule
\multicolumn{4}{c}{\textbf{General multi-turn instruction-following \& consistency}} \\
\midrule

\sys{MT-Bench}~\cite{zheng2023judging}
  & T & 2     & J   \\
\sys{MT-Bench-101}~\cite{bai-etal-2024-mt}
  & T & 3     & J   \\

\sys{MT-Eval}~\cite{kwan2024mteval}
  & T & 7     & M   \\
\sys{MultiChallenge}~\cite{deshpande-etal-2025-multichallenge}
  & T & $5$ avg / $\leq10$ & M   \\

\sys{IHEval}~\cite{zhang2025iheval}
  & T & 2 & R   \\
\sys{TURNWISE}~\cite{graf2026turnwise}
  & T & $2$--$8$ & J   \\
\sys{TOD-ProcBench}~\cite{ghazarian2025tod}
  & T & -- & R   \\
\sys{EvolIF}~\cite{jia2025one}
  & T & -- & M   \\
\sys{Parrot-Bench}~\cite{sun2024parrot}
  & T & 8 & J   \\
\sys{$\tau$-Bench}~\cite{yao2025tau}
  & T & -- & M   \\
\sys{StructFlowBench}~\cite{li2025structflowbench}
  & T & 4.14 avg & J   \\
\sys{PersonaMem}~\cite{jiang2025know}
  & T & $\leq$60 sessions& Ac \\

\sys{CORAL}~\cite{cheng-etal-2025-coral}
  & T & 8.26 avg & M   \\
\sys{ToolSandbox}~\cite{lu-etal-2025-toolsandbox}
  & T & 13.9 avg & rM   \\
\sys{TurnBench-MS}~\cite{zhang-etal-2025-turnbench}
  & T & >20  & M   \\

\sys{MINT}~\cite{wang2024mint}
  & T & $\leq$5 & R  \\
\sys{SOTOPIA}~\cite{zhou2024sotopia}
 & T & $\leq$20 & M \\

\sys{AgentBoard}~\cite{ma2024agentboard}
  & T & 3--25 & M   \\
\sys{MemoryAgentBench}~\cite{hu2025evaluating}
& T & -- & M \\

\midrule
\multicolumn{4}{c}{\textbf{Multilingual \& cross-lingual}} \\
\midrule

\sys{M2Lingual}~\cite{maheshwary2025m2lingual}
  & T & 2  & J   \\
\sys{CMT-Eval}~\cite{tian2025cmt}
  & T & multi & J   \\

\sys{AlignMMBench}~\cite{wu2025alignmmbench}
  & T+V & -- avg   & M   \\
\sys{MT-Bench-Hi}~\cite{kamath-etal-2025-benchmarking}
  & T & 2 & J   \\
\sys{cuDialog}~\cite{cao-etal-2024-bridging}
  & T & $8$ sent. $(5{+}3)$& R  \\
\sys{IndoToD}~\cite{kautsar-etal-2023-indotod}
  & T & $2.63/4.06$ avg & R   \\

\midrule
\multicolumn{4}{c}{\textbf{Multimodal (image + text)}} \\
\midrule

\sys{ConvBench}~\cite{liu2024convbench}
  & T+V & 3 & J  \\
\sys{MMDU}~\cite{liu2024mmdu}
  & T+V & $15$ avg / $27$ max    & J  \\

\sys{MMCR}~\cite{yan2025mmcr}
  & T+V & $4$/$8$ & J \\
\sys{MMRC}~\cite{xue2025mmrc}
  & T+V & $\$15.2$ avg & M \\

\sys{MultiVerse}~\cite{lee2025multiverse}
  & T+V & $3.91$ avg  & J \\
\sys{Mem-Gallery}~\cite{bei2026mem}
  & T+V & MS    & M  \\
\sys{MMMT-IF}~\cite{epstein2024mmmtif}
  & T+V & 1--20 & M  \\
\sys{MMMB}~\cite{tong2025interactiveomni} & T+V & $\leq$15 & J \\
\midrule
\multicolumn{4}{c}{\textbf{Spoken \& video}} \\
\midrule

\sys{Audio MultiChallenge}~\cite{gosai2025audiomultichallenge}
  & S     & $3$--$8$ & M  \\
\sys{MT-Video-Bench}~\cite{pan2025mtvidobench}
  & T+Vid & $5$--$8$ & J \\

\sys{OmniMMI}~\cite{wang2025omnimmi}
  & T+S+Vid & stream / $1$--$3$ & M \\
\sys{SCVBench}~\cite{ijcai2025p255}
  & T+Vid & $6.16$ avg & R  \\

\sys{CogStream}~\cite{zhao2026cogstream}
  & T+Vid &  stream; $5.02$ & J  \\
\sys{AVHBench}~\cite{sung-bin2025avhbench}
  & T+S+V & -- & M      \\
\sys{MTalk-Bench}~\cite{du2025mtalk}
  & S & $2$--$3$ & M        \\
\sys{FD-Bench}~\cite{peng25b_interspeech}
  & S & ${\leq}5$ & M        \\
\sys{MULTI-Bench}~\cite{deng2025multibench}
  & S & 8 avg  & M      \\
\sys{SVBench}~\cite{yang2025svbench}
  & T+Vid &  stream; $4.29$ & J   \\
\sys{MSIB}~\cite{tong2025interactiveomni} & S & 2--10 & M \\
\midrule
\multicolumn{4}{c}{\textbf{Robustness, fairness \& safety}} \\
\midrule

\sys{FB-Bench}~\cite{li2025fb}
  & T & 2     & M   \\
\sys{FairMT-Bench}~\cite{ICLR2025_00d80722}
  & T & multi & M   \\

\sys{SYCON-Bench}~\cite{hong2025sycon}
  & T & multi & R   \\
\sys{Curse of Multi-Modalities}~\cite{leng2026the}
  & T+S+V & -- & M  \\
\sys{Lost in Multi-Turn}~\cite{laban2025lostmultiturn}
  & T & multi & M   \\
\sys{X-Teaming}~\cite{rahman2025x}
  & T & avg 5.10 & M \\
\sys{Crescendo}~\cite{russinovich2025great}
  & T & 1--7 & R    \\
\sys{DiaHalu}~\cite{chen-etal-2024-diahalu}
  & T & 6.91 avg & H+J \\
\sys{SafeDialBench}~\cite{cao2026safedialbench}
  & T & 3--10 & M   \\

\bottomrule
\end{tabular}}
\vspace{-0.3cm}
\caption{%
Multi-turn evaluation benchmarks grouped by modality and evaluation focus.
\textbf{Mod.} denotes modality, with T for text, S for speech, V for vision, and Vid for video.
\textbf{Turns} reports the average number of turns per instance or session when available, and otherwise gives a qualitative range.
\textbf{Eval} denotes the evaluation protocol, with R for rule-based scoring, H for human evaluation, J for LLM-as-judge, M for mixed evaluation, and Ac for accuracy.
Benchmarks that evaluate only single-turn ability are excluded.
}
\label{tab:benchmarks-mt}
\vspace{-0.35cm}
\end{table}

%% file: sections_short/modeling_and_challenges.tex
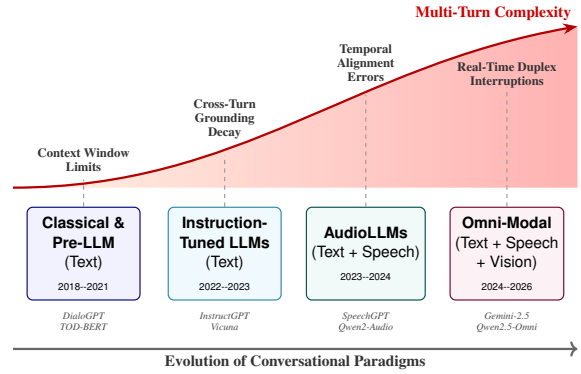
\begin{figure}[t] 
\centering
\resizebox{0.98\linewidth}{!}{%
\begin{tikzpicture}[
    x=3.5cm, y=1cm,
    every node/.style={font=\sffamily},
    erabase/.style={
        thick, rounded corners=4pt,
        minimum width=2.8cm, minimum height=2.3cm, 
        align=center, anchor=south
    },
    model/.style={font=\scriptsize\itshape, text=black!70, align=center, anchor=north},
    challenge/.style={font=\small\bfseries, text=black!80, align=center}
]

\shade[left color=orange!10, right color=red!30, draw=none] 
    (-0.5, 2.8) to[out=0, in=190] (3.5, 6.8) -- (3.5, 2.8) -- cycle;
    
\draw[ultra thick, red!70!black, -Stealth] (-0.5, 2.8) to[out=0, in=190] (3.5, 6.8)
    node[above left, font=\bfseries\normalsize, text=red!80!black] {Multi-Turn Complexity};

\node[erabase, draw=blue!50!black, fill=blue!5] (era0) at (0, 0) {
    \textbf{Classical \&}\\\textbf{Pre-LLM}\\(Text)\\[0.1cm]\scriptsize 2018--2021};
    
\node[erabase, draw=cyan!60!black, fill=cyan!5] (era1) at (1, 0) {
    \textbf{Instruction-} \\\textbf{Tuned }\textbf{LLMs}\\(Text)\\[0.1cm]\scriptsize 2022--2023};
    
\node[erabase, draw=teal!60!black, fill=teal!5] (era2) at (2, 0) {
    \vphantom{\textbf{P}}\textbf{AudioLLMs}\\\vphantom{\textbf{P}}(Text + Speech)\\[0.1cm]\scriptsize 2023--2024};
    
\node[erabase, draw=purple!60!black, fill=purple!5] (era3) at (3, 0) {
    \textbf{Omni-Modal}\\(Text + Speech \\+ Vision)\\[0.1cm]\scriptsize 2024--2026};

\node[model] at (era0.south) [yshift=-0.15cm] {DialoGPT\\TOD-BERT};
\node[model] at (era1.south) [yshift=-0.15cm] {InstructGPT\\Vicuna};
\node[model] at (era2.south) [yshift=-0.15cm] {SpeechGPT\\Qwen2-Audio};
\node[model] at (era3.south) [yshift=-0.15cm] {Gemini-2.5\\ Qwen2.5-Omni};

\draw[->, ultra thick, draw=black!60] (-0.5, -1.2) -- (3.5, -1.2)
    node[midway, below, font=\bfseries, text=black!80] {Evolution of Conversational Paradigms};

\node[challenge] at (0, 3.5) {Context Window\\Limits};
\node[challenge] at (1, 4.5) {Cross-Turn\\Grounding \\Decay};
\node[challenge] at (2, 5.9) {Temporal\\Alignment \\Errors};
\node[challenge] at (3, 5.6) {Real-Time Duplex\\Interruptions};

\draw[dashed, thick, black!40] (era0.north) -- (0, 3.0);
\draw[dashed, thick, black!40] (era1.north) -- (1, 3.6);
\draw[dashed, thick, black!40] (era2.north) -- (2, 5.1);
\draw[dashed, thick, black!40] (era3.north) -- (3, 5.3);

\end{tikzpicture}
} 
\vspace{-0.3cm}
\caption{Evolution of conversational architectures. Multi-turn challenges escalate as systems evolve from simple text to complex omni-modal agents. %
}
\label{fig:evolution_complexity}
\vspace{-0.35cm}
\end{figure}

\section{Modeling Paradigm}
\label{sec:modelling}

Modeling paradigms have shifted from modular dialogue pipelines to general-purpose LLMs, AudioLLMs, omni-modal systems, and tool-augmented agents, as shown in Figure~\ref{fig:evolution_complexity}. Appendix Table~\ref{tab:models-full} summarizes the details. This evolution has broadened the interaction interface, however, it has not solved the central multi-turn challenge. Most systems still treat dialogue history as a flat context, while only a smaller body of work explicitly models memory, state updates, selective grounding, and efficient long-session reasoning.

\noindent\textbf{Classical systems} separated language understanding, dialogue state tracking, policy learning, and response generation, often using rule-based, statistical, or neural sequence-to-sequence components.

\noindent\textbf{Transformer-era} dialogue models began to consolidate these functions through pretraining, multi-task learning, knowledge injection, and explicit memory mechanisms.

\noindent \textbf{Instruction-tuned LLMs} then shifted dialogue modeling away from task-specific managers toward general-purpose models trained or adapted for instruction following, open-ended interaction, and structured task behavior through prompting or function-like abstractions. %
The next stage added speech, vision, and real-time interaction.

\noindent\textbf{AudioLLMs} extend conversational AI from text-only interaction to spoken dialogue by modeling speech alongside language. Early systems represent speech as discrete tokens within an LLM framework, enabling speech input and output to be handled as part of the same sequence modeling problem. Later models broaden this direction toward chat-ready audio understanding, generic hearing, and interleaved speech-text modeling.

A second line targets real-time spoken interaction, supporting streaming generation, controllable voice, emotion, timbre, and full-duplex dialogue with low latency. Textless spoken dialogue models further show that dialogue can be modeled directly from raw audio, preserving turn-taking and paralinguistic cues often lost in transcripts.

Overall, AudioLLMs move dialogue systems toward speech-native interaction. Yet most still handle multi-turn context through the underlying language model, with limited explicit modeling of long-horizon spoken memory, interruptions, and cross-turn acoustic grounding.

\noindent \textbf{Omni-modal} models extend AudioLLMs by jointly handling text, speech, vision, and sometimes video. Recent systems explore several design choices, including streaming duplex interaction, low-latency speech and vision alignment, shared multimodal tokenization, modality-specific encoders, and separate modules for reasoning and speech generation. These designs move conversational AI from speech-text interaction toward unified perception and generation across modalities.

\noindent\textbf{Audio-visual dialogue} forms a more specialized direction in which visual cues support speaker tracking, turn-taking, and grounding rather than serving only as general perceptual input. AV-Dialog uses lip-centered visual features to identify the target speaker, predict turn transitions, and generate responses under noise and competing speech~\cite{chen2026av}. MAViD instead combines multimodal understanding with synchronized audio-video response generation through a Conductor--Creator architecture, though it is not evaluated evaluated for sustained multi-turn interaction~\cite{pang2025mavid}. Despite this progress, most omni-modal systems still treat dialogue history as a flat sequence. Multi-turn-aware methods address grounded memory, context management, and long-session inference cost, while long-horizon grounding, reference tracking, and session-level coherence remain open challenges.

\noindent\textbf{Agentic} dialogue systems extend multi-turn modeling by framing conversation as a loop of planning, tool use, observation, and revision. ReAct interleaves reasoning and actions \cite{yao2022react}, Reflexion uses feedback and episodic memory to improve later attempts \cite{shinn2023reflexion}, and MetaGPT enables role-based collaboration through structured workflows \cite{hong2024metagpt}. Recent systems apply these ideas to tool use, function calling, web navigation, recommendation, video reasoning, and live multimodal interaction \cite{wu2024toolplanner,pmlr-v235-lu24e}. These settings introduce three key challenges. First, agents must track external states and dependencies that may not appear in the dialogue history; ToolSandbox evaluates such stateful and implicit dependencies. Second, tool calls can produce undesirable or irreversible effects, motivating evaluation of both task completion and harmful side effects. Third, failed actions require replanning and backtracking. Reflexion and SCoRe support self-correction, but do not provide transactional rollback for already executed external actions. Overall, modeling has progressed from modular state tracking to unified, multimodal, and agentic systems, while long-horizon grounding and cross-turn memory remain key open challenges. Appendix~\ref{app:modeling-paradigms} provides additional model-level details.

\section{Training Strategies}
\label{sec:training}

\input{tables/comparison_training_strats}

Multi-turn dialogue training must account for dependencies across turns, delayed rewards, changing user intent, and context-sensitive response quality. We organize existing approaches into two complementary groups. \textit{Optimization and supervision strategies} include \textit{(i)} supervised fine-tuning, \textit{(ii)} multi-task learning, and \textit{(iii)} RL and preference optimization. \textit{Data construction and grounding strategies} include \textit{(iv)} synthetic data generation and \textit{(v)} conversational retrieval-augmented training. Table~\ref{tab:train-compare} provides a conceptual comparison of these strategies in terms of their learning mechanisms, cross-turn roles, benefits, and main limitations. Below we briefly discuss each approach.

\noindent\textbf{\textit{(i)} Supervised fine-tuning.}
Supervised fine-tuning requires conversations that reflect real multi-turn phenomena such as follow-up questions, anaphora, ellipsis, topic shifts, and safety escalation. UltraChat and WildChat provide large-scale synthetic and real multi-turn chat data, while Parrot explicitly targets referential phenomena in follow-up turns. Domain-specific resources such as Aquila-Med, Qilin-Med, and Zhongjing adapt SFT to multi-turn clinical dialogue, and XGuard-Train extends SFT to multi-turn %
safety training.

\noindent\textbf{\textit{(ii)} Multi-task learning.}
Multi-task learning improves transfer by sharing representations across dialogue objectives. PPTOD jointly trains response generation, dialogue state tracking, and policy prediction, while TOD-BERT learns dialogue-aware representations across task-oriented dialogue corpora. DAMSEL shows that auxiliary objectives are useful for conversational QA when labeled speech data is limited.

\noindent\textbf{(\textit{iii)} RL and preference optimization.}
Multi-turn reinforcement learning shifts the learning signal from isolated responses to full interactions. InstructGPT establishes the standard RLHF pipeline, while ArCHer and MT-RLHF adapt optimization to longer conversations and conversation-level preferences. Preference methods such as DMPO, Multi-turn DPO/KTO, Parrot, and SDPO optimize preferences at response, segment, or trajectory level. Other work trains action selection, tool use, self-correction, proactive interaction, and long-horizon sparse-reward behavior across turns.
Clinical and tutoring settings further combine SFT, RLHF, DPO, expert feedback, and multi-agent RL for specialized multi-turn interaction.

\noindent\textbf{\textit{(iv)} Synthetic data generation.}
Synthetic generation is widely used to scale multi-turn data and control dialogue structure. UltraChat, MMDU-45K, MMDiag, and TMDialog generate instruction-following, multimodal, note-taking, and context-modeling dialogues. Other pipelines target Arabic dialogue, emotional speech, tool use, coherent intent trajectories, multi-topic information seeking, reviewer-guided regeneration, task-oriented bootstrapping, and multi-turn API/tool-use simulation.

\noindent\textbf{\textit{(v)} Conversational RAG.}
Conversational RAG trains models to retrieve, filter, and use evidence across turns. PK-ICR and commonsense retrieval approaches ground responses in persona, external knowledge, and knowledge graphs. ChatQA, ChatQA-2, HAConvDR, UniConv, IterCQR, and related work improve conversational retrieval, query reformulation, history selection, and joint retrieval-generation training. KEDiT and CORAL further explore efficient evidence integration and citation-aware conversational RAG training. To translate this comparison into practice, in Table~\ref{tab:decision}, we map common deployment goals to the training strategy. In Appendix~\ref{app:modeling-paradigms}, we provide additional details on training strategies.

Overall, these approaches address complementary aspects of multi-turn learning: supervised and multi-task objectives support history-conditioned behavior and intermediate dialogue structure; reinforcement learning and preference optimization target interaction-level outcomes; synthetic data expands trajectory coverage; and conversational RAG improves grounding across turns. In the appendix, Table~\ref{tab:models-full} summarizes the corresponding modeling paradigms, Table~\ref{tab:training-full} provides a detailed overview of representative training methods and systems, and Table~\ref{tab:decision} translates these findings into practical guidance by mapping common deployment goals to suitable training strategies.

%% file: tables/comparison_training_strats.tex
\begin{table*}[t]
\centering

\small
\setlength{\tabcolsep}{2.5pt}
\scalebox{0.95}{
\begin{tabularx}{\textwidth}{@{}>{\raggedright\arraybackslash}p{2.7cm}>{\raggedright\arraybackslash}p{3.25cm}>{\raggedright\arraybackslash}p{3.0cm}>{\raggedright\arraybackslash}X>{\raggedright\arraybackslash}X@{}}
\toprule
\textbf{Strategy} & \textbf{Learning mechanism} & \textbf{Cross-turn role} & \textbf{Primary benefit} & \textbf{Main limitation} \\
\midrule
\multicolumn{5}{@{}l}{\textbf{Optimization and supervision strategies}} \\
\addlinespace[2pt]
\textbf{Supervised fine-tuning} & Next-token likelihood over reference multi-turn dialogues & History-conditioned response generation & Provides a stable foundation for history-conditioned dialogue and domain adaptation & Provides no explicit credit for delayed consequences and inherits biases in the demonstrations \\
\addlinespace
\textbf{Multi-task learning} & Dialogue loss combined with auxiliary objectives for state, persona, knowledge, consistency, retrieval, or policy & Explicit supervision of intermediate states, consistency, and grounding & Improves intermediate-state supervision, grounding, and transfer across dialogue skills & Requires auxiliary labels and careful loss balancing; unrelated objectives may cause negative transfer \\
\addlinespace
\textbf{RL and preference optimization} & Rewards or preference judgments over responses, dialogue segments, or complete trajectories & Interaction-level outcome optimization and delayed credit assignment & Optimizes interaction-level outcomes and enables delayed credit assignment & Response-level feedback may overlook long-horizon effects, while trajectory-level feedback is costly and noisy \\
\midrule
\multicolumn{5}{@{}l}{\textbf{Data construction and grounding strategies}} \\
\addlinespace[2pt]
\textbf{synthetic data generation} & Generated, self-play, corrected, or distilled dialogue trajectories & Coverage of long, rare, difficult, or controlled interactions & Expands the coverage of long, rare, and controlled dialogue trajectories & May propagate generator errors and biases and therefore requires validation and filtering \\
\addlinespace
\textbf{Conversational RAG} & Supervision for history-aware retrieval, evidence selection, and grounded response generation & Evidence tracking, selection, and use across turns & Grounds responses in external and updatable evidence across turns & Depends on retrieval quality and increases training and inference complexity \\
\bottomrule
\end{tabularx}
}
\vspace{-0.2cm}
\caption{Comparison of the \textbf{five training-strategy families}, summarizing each strategy's learning mechanism, cross-turn role, primary benefit, and main limitation.}
\label{tab:train-compare}
\vspace{-0.3cm}
\end{table*}

%% file: sections_short/evaluation.tex
\section{Evaluation of Multi-turn Dialogue}
\label{sec:evaluation}

Evaluating multi-turn dialogue requires moving beyond single-response metrics. A response may be fluent and factually correct on its own; however, it can still fail at the session level if it ignores earlier instructions, loses user preferences, misuses retrieved evidence, mishandles tool state, or breaks grounding across modalities. Evaluation therefore needs to measure both local response quality and cross-turn consistency, memory, state tracking, grounding, interaction success, and robustness across the full dialogue.

Existing evaluation methods fall into five broad groups. 
\textit{(i)} Surface-form and task-oriented metrics, such as overlap scores, semantic similarity, dialogue state tracking, and task success, remain useful, though they provide limited insight into session-level coherence. 
\textit{(ii)} Session-level metrics assess instruction retention, memory recall, constraint satisfaction, feedback integration, and dialogue-level hallucination. 
\textit{(iii)} Agentic and retrieval-grounded evaluation measures tool use, stateful execution, evidence retrieval, and source attribution. 
\textit{(iv)} Speech-native and full-duplex evaluation adds speech quality, paralinguistic cues, interruption handling, latency, and response timing. 
\textit{(v)} LLM-as-a-judge and human evaluation support open-ended assessment; however, their reliability depends on rubric quality and may be affected by judge bias, rater inconsistency, and small ranking differences. 
In Table~\ref{tab:eval-compare}, we compare these five families by what they measure, the systems they best suit, and their main limitations, showing that no single family fully captures session-level competence. Appendix~\ref{app:evaluation-details} provides a detailed discussion of these approaches, and Table~\ref{tab:eval-metrics-full} summarizes representative metrics and frameworks.

\noindent\textbf{Evaluation gaps.}
Current setups make multi-turn failures more visible, however, several gaps remain. 
\textit{(i)} Strong single-turn performance does not reliably transfer to session-level interaction. 
\textit{(ii)} Agentic dialogue lacks unified evaluation for hidden state, tool-side effects, source attribution, and repeated-trial reliability. 
\textit{(iii)} Spoken and full-duplex systems still lack integrated session-level evaluation that jointly measures semantic correctness, audio quality, timing, interruptions, and paralinguistic behaviour. 
\textit{(iv)} Multi-turn evaluation remains focused in English and high-resource settings, leaving multilingual, dialectal, and culturally grounded evaluation split across separate resources. These gaps show that current evaluation still measures many components of dialogue competence separately, rather than testing reliable session-level behaviour across languages, modalities, tools, and time.

%% file: sections_short/challenges_gaps.tex
\section{Challenges, and %
Future Directions}
\label{sec:challenges-future}

The studied literature shows progress in datasets, modeling, training, and evaluation, however, sustained multi-turn interaction remains unresolved. Stronger single-turn models alone do not achieve multi-turn competence. Systems must preserve context, revise assumptions, ground responses, coordinate tools, and maintain quality across long, multimodal, and culturally diverse sessions. Mechanisms for memory, cross-turn grounding, speech-native interaction, robust revision, and session-level evaluation remain limited.

\paragraph{Persistent memory and state management.}
Current long-context models can process more dialogue history, however, they still struggle to retrieve, update, and apply the right information across turns. This problem grows in multi-session settings, where preferences, facts, and task states change over time. Future work should move beyond flat context windows toward explicit memory systems that separate dialogue state, 
and task-specific working memory. These systems should support updates, forgetting, conflict resolution, and transparent retrieval across sessions.

\paragraph{Cross-turn grounding.}
Grounding weakens when responses depend on earlier turns, external evidence, visual content, acoustic cues, or tool outputs. Full dialogue history can add noise in retrieval-augmented dialogue, while multimodal references often decay across turns. Future work should develop selective grounding methods that identify the relevant dialogue history, evidence, and modality streams for each turn. This requires structured dialogue state rather than treating the full history as a flat input sequence.

\paragraph{Speech-native and full-duplex interaction.}
Spoken dialogue adds timing, prosody, interruptions, repairs, noise, and paralinguistic cues that transcript-only pipelines often miss. Full-duplex systems make this harder because users and systems may speak at the same time. Future work should treat speech as an interaction medium, not only as an input modality. This requires training and evaluation for interruption handling, response timing, spoken clarification, voice-grounded tool use, and paralinguistic understanding.

\paragraph{Robustness and revision.}
Multi-turn systems often commit to early assumptions and fail to revise them after later corrections. They are also vulnerable to adversarial escalation, where individually benign turns accumulate into unsafe or incorrect outcomes. Future work should improve session-level robustness by enabling models to detect uncertainty, recover from mistakes, and maintain safety throughout the interaction.

\paragraph{Evaluation mismatch.}
Current evaluation methods capture only parts of multi-turn behavior. Surface-form metrics, task-oriented metrics, agentic benchmarks, speech-native evaluation settings, and LLM-as-judge methods remain difficult to compare. No unified framework yet measures session-level competence across memory, grounding, tool use, speech, safety, and user satisfaction. Future evaluation should move from turn-level scoring to session-level assessment, with explicit measures of state consistency, evidence use, revision, timing, and repeated-trial reliability.

\paragraph{Cultural and linguistic coverage.}
Most resources and evaluations still focus on English and high-resource settings. Dialects, code-switching, cultural pragmatics, and region-specific knowledge shape how users express intent and judge appropriate responses. Future benchmarks should move beyond isolated language-specific datasets toward comparable multilingual and culturally grounded evaluation suites. This is especially important for %
multimodal settings, where language, culture, accent, and visual context interact.

\paragraph{Beyond turn-by-turn exchange.}
Most systems still follow a simple user-query and assistant-response loop. This framing does not fully capture proactive suggestions, mid-utterance clarification, streaming perception, emotion-aware interaction, or mixed-initiative collaboration. Future systems need interaction models that treat dialogue as continuous, stateful, and adaptive rather than as a sequence of independent turns.

%% file: sections_short/appendix.tex
\section{Appendix}
\subsection{Additional Resources}
\subsubsection{Text-only Multi-turn Resources}
\label{app:text-mt-resources}

\paragraph{Foundational benchmarks.}
CoQA~\cite{reddy2019coqa} contains conversational question answering grounded in source passages and emphasizes dependencies across previous turns. MultiWOZ~2.1~\cite{eric2020multiwoz} provides multi-domain task-oriented dialogues with corrected state annotations and is widely used for dialogue state tracking, inform rate, and task success. PersonaChat~\cite{zhang2018personachat} pairs open-domain conversations with explicit persona profiles, making it a standard resource for persona-consistency research.

\paragraph{Capability evaluation.}
MT-Bench-101~\cite{bai-etal-2024-mt} introduces a hierarchical taxonomy for fine-grained multi-turn evaluation. MT-Eval~\cite{kwan2024mteval} evaluates recollection, expansion, refinement, and follow-up, showing that models can degrade when a task is distributed across turns. MultiChallenge~\cite{deshpande-etal-2025-multichallenge} targets realistic human-LLM conversations and evaluates instruction retention, inference memory, reliable versioned editing, and self-coherence. MINT~\cite{wang2024mint} restructures reasoning, code, and decision-making tasks into a user-tool-LLM interaction format.

\paragraph{Long-term memory.}
LoCoMo~\cite{maharana2024evaluating} evaluates long-term conversational memory through extended multi-session dialogues grounded in personas and temporal event graphs. LongMemEval~\cite{wu2025longmemeval} tests information extraction, multi-session reasoning, knowledge updates, temporal reasoning, and abstention. DialSim~\cite{kim2024dialsim} evaluates long-term multi-party dialogue using TV-series transcripts and response-time constraints. PersonaMem~\cite{jiang2025know} evaluates dynamic user profiling and preference tracking across multi-session histories. Mem-Gallery~\cite{bei2026mem} extends long-memory evaluation to multimodal settings.

\paragraph{Training corpora.}
UltraChat~\cite{ding2023ultrachat} provides synthetic multi-turn instruction-following conversations. WildChat~\cite{zhao2024wildchat} contains opt-in real ChatGPT conversations with broad topic and language diversity. LMSYS-Chat-1M~\cite{zheng2024lmsyschatm} collects real user conversations with multiple LLMs through public chat interfaces. DocTalk~\cite{lee-etal-2025-doctalk} synthesizes multi-turn information-seeking dialogues from Wikipedia and is mainly designed as a pre-training resource.

\paragraph{Persona, role-play, and social interaction.}
PIPPA~\cite{gosling2023pippa} contains long role-play conversations and supports evaluation of in-character consistency. PRODIGy~\cite{occhipinti2024prodigy} grounds personas in fictional character profiles and studies the effect of backstory and speaking style. SOTOPIA~\cite{zhou2024sotopia} evaluates social intelligence through role-play scenarios where agents pursue social goals while maintaining appropriate behavior.

\paragraph{Tool use, retrieval, and proactive assistance.}
$\tau$-Bench~\cite{yao2025tau} evaluates agents in stateful user-tool interaction settings with task and policy constraints. ToolSandbox~\cite{lu-etal-2025-toolsandbox} evaluates stateful tool use and interaction consistency across turns. mtRAG~\cite{katsis-etal-2025-mt} evaluates multi-turn retrieval-augmented generation, where the agent must decide when and what to retrieve as the conversation evolves. ProMISe~\cite{butala-etal-2024-promise} evaluates proactive information seeking by asking agents to generate useful follow-up suggestions during a dialogue.

\subsubsection{Spoken Dialogue Resources}
\label{app:spoken-mt-resources}

\paragraph{Foundational resources.}
TurnGPT~\cite{ekstedt-skantze-2020-turngpt} applies a Transformer language model to turn-taking prediction in spoken dialogue. It frames turn completion as a language-modeling problem over transcribed conversational text, providing an early trainable baseline for predicting when a speaker's turn is likely to end. SpokenWOZ~\cite{si2023spokenwoz} provides 249 hours of human-human task-oriented speech across eight domains and 203K turns. It preserves audio, transcripts, and dialogue-state annotations, allowing evaluation of how recognition errors affect downstream state tracking and task success.

\paragraph{Multi-turn speech-to-speech evaluation.}
Audio MultiChallenge~\cite{gosai2025audiomultichallenge} extends multi-turn evaluation to spoken dialogue systems using 452 conversations and 1,712 instance-level rubrics. It evaluates the four axes of MultiChallenge together with a Voice Editing axis for mid-utterance speech repairs, and compares turn-level and conversation-level scores. URO-Bench~\cite{yan-etal-2025-uro} evaluates end-to-end speech-to-speech dialogue models across multilingual, multi-round, and paralinguistic settings, with basic and pro tracks covering 20 S2S task types. MTalk-Bench~\cite{du2025mtalk} constructs multi-turn speech-to-speech test dialogues through an LLM-human pipeline, then adds human recording, voice conversion, and ambient sound mixing. It evaluates semantic content, vocal cues, and background sound using both arena-style pairwise comparison and rubric-based absolute scoring. MULTI-Bench~\cite{deng2025multibench} focuses on emotional intelligence in spoken dialogue and tests whether models can track and respond to emotional-state changes across turns. FD-Bench~\cite{peng25b_interspeech} evaluates simultaneous bidirectional speech through simulated five-round conversations, measuring interruption handling and response timing with metrics such as SIR, SRIR, EIR, NIR, IRD, FSED, ERT, and EIT, as referred in Table \ref{tab:eval-metrics-full}.

\paragraph{Related audio-agent benchmarks.}
ADU-Bench~\cite{gao-etal-2025-benchmarking} provides open-ended audio dialogue tasks covering multiple scenarios, skills, languages, and ambiguity types such as intonation and homophones. Because it is not explicitly multi-turn in the current categorization, it is best treated as a related audio-dialogue benchmark rather than a core multi-turn resource. Full-Duplex-Bench-v3~\cite{lin2026fdb_v3} extends full-duplex benchmarking to tool-use scenarios, measuring whether voice agents invoke tools correctly while handling disfluencies such as false starts and repairs. Since it is marked as work in progress, it can be mentioned as an emerging direction rather than a central benchmark.

\paragraph{Training corpora and specialized spoken resources.}
DeepDialogue~\cite{deepdialogue2025} provides synthetic multi-turn dialogues paired with emotionally consistent speech, making it useful for training spoken dialogue models with emotion-control signals. ASK-QA~\cite{chen-etal-2025-data} contains spoken conversational QA sessions with multiple speakers and deliberately ambiguous user requests, supporting research on clarification and spoken question answering. The MENASpeechBank~\cite{ali2026menaspeechbankreferencevoicebank} provides persona-conditioned conversations in Arabic and regional dialects across the MENA region, combining voice-bank design with multi-turn spoken Arabic dialogue.

\subsubsection{Multimodal Dialogue Resources}
\label{app:multimodal-mt-resources}

\paragraph{Foundational and open-domain resources.}
VisDial~\cite{das2017visual} pairs COCO images with ten rounds of question answering and established the standard format for multi-turn visual dialogue. MMDialog~\cite{feng2023mmdialog} provides large-scale real-world multimodal conversations with images across thousands of topics and introduces MM-Relevance for modality-aware response evaluation. DialogCC~\cite{lee2024dialogcc} and IMAD~\cite{moskvoretskii2024imad} construct image-augmented dialogues by matching images to existing text conversations using vision-language similarity, making them scalable resources for multimodal dialogue training. InfoVisDial~\cite{wen2023infovisdial} adds Wikipedia-grounded knowledge to visual dialogue, requiring models to combine visual evidence with external factual information across turns.

\paragraph{Instruction-tuning corpora.}
MMDU-45K~\cite{liu2024mmdu} contains multi-turn, multi-image dialogues generated with GPT-4o and human refinement, with sessions that can include many images and long image-text contexts. MMCR~\cite{yan2025mmcr} provides both an instruction-tuning corpus and a diagnostic benchmark covering multiple domains and subtopics, supporting evaluation under single-image and multi-image conversational settings.

\paragraph{Capability evaluation.}
ConvBench~\cite{liu2024convbench} evaluates multi-turn LVLM dialogues across perception, reasoning, and creation tasks, showing that early visual perception errors can propagate into later turns. MultiVerse~\cite{lee2025multiverse} evaluates multi-turn multimodal dialogue across many task types and explicitly tests in-context learning during conversation. MMRC~\cite{xue2025mmrc} evaluates information extraction, multi-turn reasoning, information update, image management, memory recall, and refusal behavior, identifying recurring failures such as memory degradation, update failure, and error propagation. MMMT-IF~\cite{epstein2024mmmtif} focuses on multi-turn instruction following with image inputs, making it a multimodal counterpart to text-only instruction-retention benchmarks. Chatterbox/CB-300K~\cite{tian2025chatterbox} targets multi-round referring and grounding, with emphasis on referential ambiguity, spatial relations, and grounding consistency across turns.

\paragraph{Long-term memory and context management.}
Mem-Gallery~\cite{bei2026mem} evaluates multi-session multimodal memory across visual and textual histories, including extraction, reasoning, and knowledge management. DiagNote/MMDiag~\cite{liu2025diagnote} provides multi-turn dialogues with explicit note-taking annotations and studies whether learned notes can help VLMs maintain grounded context over long interactions.

\paragraph{Emerging tri-modal, generative, cultural, and video resources.}
InteractiveOmni~\cite{tong2025interactiveomni} extends multi-turn dialogue to tri-modal audio-visual-language interaction, using cross-modal context and historical memory. DialogGen/DialogBen~\cite{huang-etal-2025-dialoggen} evaluates dialogue-conditioned image generation and editing, where the model must preserve both conversational context and output-modality coherence. AlignMMBench~\cite{wu2025alignmmbench} evaluates cultural alignment and factual consistency in single- and multi-turn Chinese and bilingual multimodal dialogues. SVBench~\cite{yang2025svbench} introduces temporal multi-turn dialogue over streaming video, where later questions depend on both earlier answers and specific video segments.

\subsubsection{Cultural and Linguistic Resources}
\label{app:datasets-cultural}
\input{tables/multi-lingual-cultural_table}

\paragraph{Arabic and dialectal multi-turn resources.}
Shawarma Chats~\cite{zeinalipour-etal-2025-shawarma} provides 30K six-turn Arabic conversations grounded in Wikipedia content, covering MSA, Egyptian Arabic, and Maghrebi Arabic. It is one of the few large-scale multi-turn resources with explicit coverage of Maghrebi Arabic, making it useful for dialect-aware dialogue modeling and evaluation. \citet{naous2020empathy} introduced an empathy-driven Arabic conversational chatbot and an accompanying message-response corpus spanning Levantine, Egyptian, and Gulf Arabic, providing an early resource for Arabic dialectal conversation modeling. Alexandria~\cite{MMAA2026} is a multi-turn conversational machine translation dataset for English-Dialectal Arabic, covering 13 Arab countries and 11 domains with city-level dialect metadata and speaker-addressee gender information. It supports evaluation of dialect naturalness, style and formality, semantic accuracy, and gender-sensitive translation.

\paragraph{Cross-lingual and culturally grounded multi-turn benchmarks.}
C3~\cite{ma-etal-2025-c3} is a bilingual English-Chinese spoken dialogue benchmark targeting phonological ambiguity, coreference, omission, and multi-turn dependency. \citet{kamath-etal-2025-benchmarking} introduce a Hindi evaluation suite consisting of MT-Bench-Hi for multi-turn instruction following, ChatRAG-Hi for conversational retrieval-augmented generation, and BFCL-Hi for function calling. cuDialog~\cite{cao-etal-2024-bridging} conditions response prediction on Hofstede cultural-value dimensions, making cultural behavior an explicit variable in multi-turn dialogue evaluation. IndoToD~\cite{kautsar-etal-2023-indotod} provides an Indonesian multi-domain task-oriented dialogue benchmark with native-speaker translations of CamRest and SMD, covering NLU, dialogue state tracking, and response generation.

\paragraph{Single-turn cultural and linguistic anchors.}
OASIS, developed through the EverydayMMQA framework~\cite{alam2025everydaymmqa}, is a culturally grounded spoken visual QA resource covering English, MSA, and Arabic dialectal varieties across 18 Arab countries. It combines real images, text questions, spoken questions, and image-grounded answers, but its current task format is single-turn rather than conversational. M2CQA~\cite{mousi2026m2cqa} provides a single-turn multimodal QA benchmark spanning 17 MENA countries in MSA and multiple Arabic dialects, and introduces the Cultural Hallucination and Factual Recall metric for measuring culturally grounded correctness. ArabicaQA~\cite{abdallah2024arabicaqa} supports Arabic question answering, Dallah~\cite{alwajih2024dallah} focuses on dialect-aware Arabic multimodal modeling, MLQA~\cite{lewis-etal-2020-mlqa} provides a cross-lingual extractive QA baseline, and mSTEB~\cite{beyene2025msteb} extends multilingual evaluation to speech and text tasks.

\paragraph{Broader cultural multimodal resources.}
SEA-VQA~\cite{urailertprasert-etal-2024-sea} evaluates cultural visual question answering across Southeast Asian settings. CVQA~\cite{mogrovejo2024cvqa}, CulturalGround~\cite{nyandwi2025grounding}, and MMA-ASIA~\cite{weihua2025mmaasia} extend culturally focused multimodal evaluation across broader language and regional settings. These resources are mostly single-turn, but they provide useful baselines and design signals for future culturally grounded multi-turn dialogue benchmarks.

\subsection{Modeling Paradigms and Training Strategies}
\label{app:modeling-paradigms}
\input{tables/modelling-architecture_table.tex}

\input{tables/training_strategies_tables}

\subsubsection{Classical Dialogue Systems}
\label{app:classical-dialogue}

Classical dialogue systems treated dialogue as a pipeline of separate modules: natural language understanding, dialogue state tracking, policy learning, and natural language generation. Early rule-based systems used hand-written templates and finite-state managers. Statistical approaches later modeled dialogue as a partially observable Markov decision process~\cite{gasic-etal-2013-pomdp}. Neural sequence-to-sequence models~\cite{sutskever2014sequence,serban2016building} replaced parts of the response-generation pipeline with end-to-end models, but the broader system design often remained modular. These systems remain useful baselines for understanding how later Transformer- and LLM-based dialogue models absorbed functions that were previously handled by separate components.

\subsubsection{Transformer Dialogue Architectures}
\label{app:transformer-dialogue}

Pre-LLM Transformer dialogue architectures established many capabilities that were later subsumed by LLMs. Open-domain systems such as DialoGPT~\cite{zhang-etal-2020-dialogpt}, BlenderBot~\cite{roller-etal-2021-recipes}, and DLGNet~\cite{oluwatobi-mueller-2020-dlgnet} improved response generation by using pretrained or contextual dialogue encoders. In task-oriented dialogue, TOD-BERT~\cite{wu2020tod}, PPTOD~\cite{su2022multi}, and the shared BERT encoder proposed by \citet{kapelonis22_interspeech} advanced unified and multi-task transformer modeling for intent recognition, slot filling, dialogue state tracking, policy prediction, and response generation.
A related line incorporated structured knowledge and dialogue state tracking. \citet{lin2021knowledge} inject schema and knowledge-graph structure into GPT-2 for dialogue state tracking, while \citet{feng-etal-2023-towards} show that LLMs can directly track slot-value pairs across turns without a dedicated DST module. Another line addresses long context and memory. Transformer-XL~\cite{dai2019transformerxl} reuses hidden states across segments, Recurrent Memory Transformer~\cite{bulatov2022recurrent} adds recurrent memory tokens, RWKV~\cite{peng2023rwkv} reformulates attention as a linear recurrence, and MemBART~\cite{membart2024} adds a gated memory module to preserve dialogue history without simply enlarging the input window.

\subsubsection{LLM-based Dialogue Models}
\label{app:llm-dialogue}

The LLM-based phase shifted dialogue modeling away from task-specific managers toward general-purpose instruction-following models. InstructGPT~\cite{ouyang2022training} helped define this phase through supervised fine-tuning, reward modeling from human preferences, and reinforcement learning from human feedback. FnCTOD~\cite{li2024large} shows that task-oriented structure can be recovered inside an LLM by treating each dialogue slot as a callable function and letting the model fill slot values zero-shot. LLM-centric systems for conversational QA and RAG, including ChatQA, ChatQA-2, and UniConv, are discussed in Section~\ref{sec:training}, since their primary contribution is training rather than architecture.

\subsubsection{AudioLLMs}
\label{app:audiollms}

AudioLLMs extend conversational models to speech by jointly modeling language and audio. Early systems such as SpeechGPT and AudioPaLM represent speech through discrete units or audio tokens within an LLM, while Qwen-Audio, Qwen2-Audio, SALMONN, Kimi-Audio, and SpiRit-LM broaden this direction toward audio understanding, spoken dialogue, and cross-modal learning~\cite{zhang2023speechgpt,rubenstein2023audiopalm,chu2023qwenaudio,chu2024qwen2,tang2024salmonn,kimi_audio_2024,nguyen-etal-2025-spirit}.

A second line focuses on real-time and full-duplex spoken interaction. Moshi supports low-latency full-duplex dialogue and uses time-aligned text tokens as an inner monologue before speech generation~\cite{defossez2024moshi}. Subsequent systems explore streaming speech generation, controllable voice, style and timbre modeling, and simultaneous listening and speaking~\cite{chen2025minmomultimodallargelanguage,fang2025llamaomni2,li2024styletalkerfinetuningaudiolanguage,zeng2024glm4voice,xiezhifei2024miniomni,chen2025slam}. Freeze-Omni keeps the LLM backbone frozen while training speech input and output modules, whereas SALMONN-omni jointly models user speech and system output in a codec-free architecture that supports turn-taking, barge-ins, and echo cancellation~\cite{wang2025freezeomni,yu2026salmonn}.

Other work separates duplex control from the dialogue backbone. FlexDuo uses a Speak-Listen-Idle state machine to reduce false interruptions, while FireRedChat combines personalized VAD with semantic end-of-turn detection for controllable barge-in~\cite{liao2025flexduo,chen2025fireredchat}. In a different direction, dGSLM models two-channel raw audio without text supervision, generating speech, laughter, and other paralinguistic signals~\cite{nguyen-etal-2023-generative}. \citet{lu2026survey} provide a broader review of full-duplex architectures, interaction ontologies, and decision states.

Overall, AudioLLMs have substantially improved speech-native and real-time interaction. However, most systems still rely on the underlying language model to maintain dialogue history, with limited explicit modeling of long-horizon spoken memory and cross-turn acoustic grounding.

\subsubsection{Omni-modal Models}
\label{app:omni-models}

Omni-modal models extend AudioLLMs by integrating vision with text and speech, and in some cases video. GPT-4o provides an end-to-end omni-modal system with real-time streaming, while Mini-Omni2 extends Mini-Omni with visual input~\cite{hurst2024gpt,xie2024miniomni2opensourcegpt4ovision,xiezhifei2024miniomni}. Other systems explore low-latency vision-speech interaction, simultaneous multimodal processing, unified tokenization, and multimodal pretraining~\cite{fu2026vita,chen2025emova,zhang2025stream,guo2025m2,wang-etal-2025-mio}. \citet{royer2025visionspeechmodels} directly connect speech and vision encoders without using text as an intermediate representation.

Recent models further broaden this design space through instruction tuning, multimodal generation, video-audio alignment, and long-horizon speech modeling~\cite{li2024baichuanomnitechnicalreport,ai2025ming,qwen25omni,wang2025mgm}. Audio-video models such as VideoLLaMA~2 and CAT additionally target reasoning over dynamic audio-visual content~\cite{damonlpsg2024videollama2,ye2024cat}.

A separate line focuses explicitly on multi-turn interaction. Most omni-modal models treat dialogue history as a flat input sequence, whereas newer methods introduce mechanisms for memory and context management. DiagNote uses explicit note-taking, CoLVLM follows a memory-perception-planning-execution loop, and ContextQFormer maintains a dedicated contextual memory block~\cite{liu2025diagnote,han2025contextual,lei2025contextqformer}. LoopServe and MadaKV reduce inference cost in long multimodal sessions, while DialogGen conditions image generation and editing on conversational history~\cite{li2025loopserve,li-etal-2025-madakv,huang-etal-2025-dialoggen}. 

Collectively, omni-modal models have broadened multimodal interaction; however, explicit cross-turn memory and long-horizon grounding remain less developed.

\subsubsection{Agentic and Tool-augmented Dialogue}
\label{app:agentic-dialogue}

Agentic dialogue systems use conversation as an interface for planning, tool use, observation, and revision. ReAct interleaves reasoning with external actions, Reflexion stores verbal feedback from failed attempts for later use, and MetaGPT coordinates multiple role-based LLM agents through structured messages~\cite{yao2022react,shinn2023reflexion,hong2024metagpt}.

Tool-augmented systems connect this process to external environments. ToolPlanner combines path planning with process- and outcome-level feedback, while ChatCoT integrates external tools into reasoning~\cite{wu2024toolplanner,chen2023chatcot}. $\tau$-Bench evaluates agents in retail and airline settings using APIs and simulated users, whereas ToolSandbox focuses on stateful tool use over variable-length interactions~\cite{yao2025tau,lu-etal-2025-toolsandbox}. xLAM-2 represents open models for multi-turn function calling~\cite{prabhakar2026apigenmt}.

Other systems extend agentic interaction to web navigation, recommendation, video, and streaming environments. WebLINX evaluates conversational web navigation, while SAPIENT applies 
Monte Carlo Tree Search to conversational recommendation~\cite{pmlr-v235-lu24e,du2025sapient}. \citet{zhang2024towards} incorporate users' emotional states into recommendation. VideoMind and Video-MTR support iterative reasoning over video, while IXC2.5-OmniLive combines streaming perception, multimodal long-term memory, and reasoning for live interaction~\cite{liu2026videomind,xie2025videomtrreinforcedmultiturnreasoning,zhang2024ixcomnilive}. 

Overall, agentic systems highlight three multi-turn requirements: tracking external state across actions, replanning as goals or environments change, and maintaining memory over extended interactions.

\subsubsection{Supervised Fine-tuning}
\label{app:sft}

Supervised fine-tuning adapts models to multi-turn interaction using conversations that preserve cross-turn dependencies. UltraChat provides large-scale synthetic instructional dialogues, while WildChat contains real opt-in conversations with broad topic and language coverage. Parrot explicitly targets anaphora and ellipsis in follow-up turns~\cite{ding2023ultrachat,zhao2024wildchat,sun2024parrot}. Domain-specific work extends SFT to clinical dialogue and adversarial safety~\cite{zhao2024aqulia,ye2023qilin,yang2024zhongjing,rahman2025x}. SFT provides a practical foundation for multi-turn training, with performance largely shaped by the coverage and quality of training data.

\subsubsection{Reinforcement Learning and Preference Optimization}
\label{app:training-rl}

Multi-turn reinforcement learning and preference optimization address delayed rewards and credit assignment across turns. InstructGPT established the standard RLHF pipeline, while ArCHer and MT-RLHF extend optimization toward turn- and conversation-level feedback~\cite{ouyang2022training,pmlr-v235-zhou24t,shani2024multiturn}. Preference-based methods such as DMPO, Multi-turn DPO/KTO, Parrot, and SDPO further optimize dialogue histories, interaction sequences, or segments rather than individual responses~\cite{shi2024direct,ICLR2025_40eff167,sun2024parrot,kong2025sdpo}.

A second line focuses on actions and outcomes across turns. Action-Based Contrastive Self-Training distinguishes actions such as answering and clarifying, while DiaTool-DPO and JOSH optimize tool-use behavior from interaction sequences~\cite{chen2025learning,jung-etal-2025-diatool,lattimer-etal-2025-sparse}. Other approaches target self-correction, collaborative reasoning, proactive interaction, and sparse long-horizon rewards~\cite{kumar2025training,sweetrl2025,gao2025refuel,wang2026implicit,abdulhai2025lmrl}. Domain-specific work further applies these methods to clinical and tutoring dialogue~\cite{scarlatos2025training,feng2026doctoragent}. 

More broadly, these methods extend optimization from individual responses to cross-turn behavior and long-horizon outcomes.

\subsubsection{Multi-task Learning}
\label{app:multi-task-training}

Multi-task learning improves dialogue modeling by sharing representations across related objectives. PPTOD jointly trains response generation, dialogue state tracking, and policy prediction, while TOD-BERT learns dialogue-aware representations across multiple task-oriented dialogue corpora~\cite{su2022multi,wu2020tod}. DAMSEL extends this approach to spoken conversational QA, using auxiliary objectives to improve learning when labeled speech data is limited~\cite{chen-etal-2025-data}. 

Taken together, multi-task learning improves transfer across related dialogue skills, although most approaches do not explicitly optimize session-level dependencies.

\subsubsection{Synthetic Data Generation}
\label{app:synthetic-training}

Synthetic data generation scales multi-turn training data while allowing control over dialogue structure, topic, and modality. UltraChat simulates user-assistant conversations, while MMDU-45K, MMDiag, and TMDialog extend synthetic generation to multimodal dialogue, note-taking, and context modeling~\cite{ding2023ultrachat,liu2024mmdu,liu2025diagnote,lei2025contextqformer}. Other pipelines target Arabic and emotional spoken dialogue, tool use, coherent intent sequences, and multi-topic information seeking~\cite{misbah2026fine,deepdialogue2025,shim2025tooldial,chen2025consistentchat,lee-etal-2025-doctalk}. Recent approaches also use reviewer-guided regeneration, task-oriented dialogue bootstrapping, and simulated API interactions to improve data quality and task coverage~\cite{wu2025instruct,ulmer-etal-2024-bootstrapping,prabhakar2026apigenmt}. 

In summary, synthetic generation offers a scalable way to target multi-turn behaviors, shifting from general dialogue scaling toward targeted simulation across modalities and tasks.

    \subsubsection{Training for Conversational RAG}
    \label{app:training-rag}
    
    Conversational RAG trains models to retrieve and use evidence as dialogue context evolves. PK-ICR jointly retrieves persona and external knowledge, while commonsense retrieval helps fill knowledge gaps in open-domain dialogue~\cite{oh-etal-2023-pk,varshney2022commonsense}. ChatQA and ChatQA-2 jointly train retrieval and generation for conversational and long-context RAG~\cite{liu2024chatqa,xu2025chatqa}. Other methods improve history selection, retrieval supervision, and query reformulation: HAConvDR filters noisy dialogue history, UniConv jointly optimizes retrieval and generation, ConvAUG strengthens retrieval through LLM-cognition data augmentation, and IterCQR iteratively reformulates queries using retrieval feedback~\cite{mo-etal-2024-history,mo-etal-2025-uniconv,chen-etal-2024-generalizing,jang-etal-2024-itercqr}. KEDiT compresses retrieved evidence into adapter parameters, while CORAL incorporates citation labeling into conversational RAG training and evaluation~\cite{zhang-etal-2025-efficient-tuning,cheng-etal-2025-coral}.
    
    Overall, conversational RAG has progressed toward history-aware retrieval and evidence integration across turns. 
    In Table~\ref{tab:decision}, we map common multi-turn training goals to the primary strategies discussed above, providing a practical guide for strategy selection.

\begin{table}[t]
\centering\footnotesize
\setlength{\tabcolsep}{4pt}
\renewcommand{\arraystretch}{1.25}
\begin{tabular}{@{}p{3.7cm} p{3.6cm}@{}}
\toprule
\textbf{Training goal} & \textbf{Primary strategy} \\
\midrule
General conversational and instruction-following ability 
& SFT on curated multi-turn dialogues \\

Domain-specific or safety-focused behavior 
& SFT with targeted synthetic augmentation \\

Human preference and long-horizon outcomes 
& Preference optimization or multi-turn RL \\

Knowledge-intensive multi-turn QA 
& Conversational RAG training \\

Low-resource settings or rare behaviors 
& Synthetic data generation \\

Persona/knowledge grounding and skill transfer 
& Multi-task learning \\
\bottomrule
\end{tabular}
\vspace{-0.2cm}
\caption{Mapping common multi-turn training goals to their primary training strategies. These strategies can also be combined depending on the target setting.}
\label{tab:decision}
\vspace{-0.3cm}
\end{table}

\input{tables/evaluation_table}

\subsection{Details on Evaluation}
\label{app:evaluation-details}

In Table~\ref{tab:eval-metrics-full}, we summarize representative metrics and frameworks across the five evaluation families. In Table~\ref{tab:eval-compare}, we compare these families by their evaluation focus, suitable settings, and main limitations. We discuss each family in detail below.

\subsubsection{Evaluation Families}
\label{app:evaluation-families}

\paragraph{Surface-form and task-oriented metrics.}
Surface-form metrics such as BLEU, ROUGE, and BERTScore remain useful baselines; however, they mainly score individual responses and provide limited insight into cross-turn behavior. They do not capture whether a model preserves instructions, persona, grounding, or coherence across a session. Task-oriented metrics such as joint goal accuracy, Slot-F1, Inform, and Success better reflect multi-turn task completion by tracking dialogue states and outcomes. However, they remain task-specific, can be sensitive to paraphrasing, and often reduce success to binary outcomes. MM-Relevance extends response relevance to image--text dialogue, although it is limited to two-modal settings.

\paragraph{Session-level dialogue metrics.}
Session-level metrics directly assess behaviors that depend on earlier turns, including memory recall, instruction retention, constraint satisfaction, feedback integration, and dialogue-level hallucination. Representative approaches measure rubric compliance (APR and ARS), positional consistency (PWC), multimodal memory and reasoning (MMRC), interaction patterns (MT-Eval), structural constraints (WCSR), instruction hierarchy (IHEval), programmatic instruction following (PIF), feedback-based reasoning (TurnBench-MS), and dialogue-level hallucination (DiaHalu). Together, these metrics provide a broader view of cross-turn competence, although their definitions and scoring remain fragmented across benchmarks.

\paragraph{Agentic and retrieval-grounded evaluation.}
Agentic and retrieval-grounded systems require evaluation of both responses and actions across an interaction. A fluent answer can still fail if the model uses an incorrect tool state, calls an API with missing information, or cites the wrong source. CORAL jointly measures retrieval, generation, and citation attribution, while $\tau$-Bench measures repeated-trial task success through $\tau$-pass$^k$. ToolSandbox evaluates stateful interactions using milestone and minefield scoring, and AgentBoard tracks progress through subgoal completion. These settings make external state, side effects, and repeated-trial reliability central evaluation concerns.

\paragraph{Speech-native and full-duplex evaluation.}
Spoken dialogue evaluation must consider both semantic content and interaction dynamics. ADU-Bench covers open-ended audio dialogue across skills, languages, and ambiguity types, while MTalk-Bench combines pairwise and rubric-based scoring for semantic content, vocal cues, and ambient sound. ASK-QA focuses on spoken clarification, and URO-Bench separates correctness, speech quality, speech--text matching, and first-packet latency. FD-Bench adds interruption and timing measures for full-duplex interaction, while Full-Duplex-Bench-v3 extends evaluation to tool use and disfluency handling. These benchmarks show that transcript correctness alone is insufficient for evaluating spoken interaction.

\paragraph{LLM-as-a-judge and human evaluation.}
LLM-as-a-judge and human evaluation support dimensions that are difficult to score automatically, including open-ended coherence, persona consistency, multimodal grounding, and human-likeness. MT-Bench, GPTScore, JudgeLM, BotChat, ContextualJudgeBench, LLM-Eval Analysis, and Multi-Judge Evaluator represent judge-based approaches, while ABC-Eval, MMDU, MTalk-Bench, and WildBench incorporate human or hybrid assessment. These methods offer flexible evaluation, although their reliability depends on rubric design and can be affected by judge bias, rater inconsistency, cost, and small ranking differences.

Overall, these evaluation methods capture complementary aspects of multi-turn performance, but no single approach fully measures session-level competence across modalities and interaction settings.

\begin{table*}[!tbh]
\centering
\scriptsize
\setlength{\tabcolsep}{3.5pt}
\scalebox{0.95}{
\begin{tabularx}{\textwidth}{
    @{}
    p{2.0cm}
    p{4.1cm}
    p{1.7cm}
    c c c
    X
    @{}
}
\toprule
\textbf{Survey} &
\textbf{Conceptual framing} &
\textbf{Modality} &
\textbf{MT} &
\textbf{Cult.} &
\textbf{Eval$\Delta$} &
\textbf{Key distinction} \\
\midrule

\citet{yi2025survey} &
LLM-based multi-turn dialogue; methods, tasks, resources, and evaluation &
T &
\cmark & \xmark & \cmark &
Broad synthesis centered on text; no common framework for comparing session-level requirements across modalities. \\

\citet{wang2023survey} &
Language-model dialogue; evolution of task-oriented and open-domain systems &
T &
\cmark & \xmark & \xmark &
Historical view of dialogue systems across language-model generations, primarily in text. \\

\citet{zhang2025survey} &
Multi-turn LLM capabilities; instruction following, memory, planning, reasoning, and evaluation &
T &
\cmark & \xmark & \cmark &
Capability-centered analysis with limited comparison across speech, vision/video, and omni-modal interaction. \\

\citet{li2025beyond} &
Multi-turn LLM interaction; task families and model-, memory-, and agent-based methods &
T &
\cmark & \pmark & \cmark &
Broad multi-turn taxonomy; multimodal session behavior is not a central comparison dimension. \\

\citet{guan2026evaluating} &
Conversational-agent evaluation; tools, memory, planning, and interaction quality &
T + Tool &
\cmark & \xmark & \cmark &
Detailed agent-evaluation framework without joint analysis of data, models, training, and evaluation across modalities. \\

\citet{zhang2024mm} &
Multimodal LLMs; architectures, training, model taxonomy, and benchmarks &
T + V (+S) &
\xmark & \xmark & \xmark &
Broad multimodal coverage, with dialogue treated mainly as an application rather than the primary unit of analysis. \\

\midrule

\textbf{This survey} &
\textbf{Complete multi-turn session; modality breadth $\times$ session-level competence across data, models, training, and evaluation} &
\textbf{T + S + V + Vid + O + Tool} &
\cmark & \cmark & \cmark &
\textbf{Cross-modal analysis using common session-level capabilities, including memory, revision, grounding, state tracking, timing, safety, and cultural--linguistic competence.} \\

\bottomrule
\end{tabularx}
}
\vspace{-0.15cm}
\caption{Comparison with closely related surveys by conceptual framing and scope.
\textbf{Modality}: T = text, S = speech/audio, V = vision, Vid = video,
O = omni-modal, Tool = tool-augmented.
\textbf{MT}, \textbf{Cult.}, and \textbf{Eval$\Delta$} indicate multi-turn,
cultural--linguistic, and multi-turn evaluation coverage, respectively.
\pmark{} denotes partial coverage.}
\label{tab:survey-positioning-detailed}
\vspace{-0.3cm}
\end{table*}

\subsection{Search Keywords}
\label{app:keywords}

We organized the literature search into six query groups covering core multi-turn dialogue, spoken and audio interaction, multimodal and omni-modal dialogue, training and alignment, evaluation, and cultural and multilingual settings. Within each group, \texttt{OR} connected alternative keywords, while \texttt{AND} combined complementary concepts. This strategy aimed to provide broad coverage across modalities, methods, benchmarks, and language settings.

\noindent\textbf{Core multi-turn dialogue.}
We used ``multi-turn'', ``multi turn'', ``multiturn'', ``multi-round'', ``conversational AI'', ``dialogue system'', ``dialog system'', ``task-oriented dialogue'', ``task oriented dialog'', ``open-domain dialogue'', ``chit-chat'', ``conversational agent'', ``dialogue state tracking'', ``conversation history'', ``context-aware dialogue'', and ``dialogue management''. These terms were combined with ``language model'', ``LLM'', ``large language model'', ``transformer'', ``BERT'', ``GPT'', ``T5'', ``instruction tuning'', ``fine-tuning'', ``pre-trained model'', and ``neural dialogue''.

\noindent\textbf{Spoken and audio dialogue.}
Speech-related keywords included ``spoken dialogue'', ``speech dialogue'', ``voice assistant'', ``spoken language understanding'', ``SLU'', ``spoken QA'', ``end-to-end spoken dialogue'', ``speech-to-speech'', ``spoken conversational'', ``audio dialogue'', ``ASR dialogue'', ``automatic speech recognition'', ``text-to-speech'', ``TTS'', ``speech synthesis'', ``spoken language model'', and ``audio language model''. We combined these with ``multi-turn'', ``conversational'', ``dialogue'', or ``multi-round''.

\noindent\textbf{Multimodal and omni-modal dialogue.}
We used ``multimodal dialogue'', ``visual dialogue'', ``visual QA dialogue'', ``video dialogue'', ``image-grounded conversation'', ``vision-language dialogue'', ``VQA multi-turn'', ``omni-modal'', ``any-to-any'', ``speech-vision'', ``multimodal conversational agent'', ``embodied dialogue'', ``audio-visual dialogue'', ``multi-modal chat'', and ``visual chatbot''. We also included representative model families such as ``LLaVA'', ``InstructBLIP'', ``Flamingo'', ``GPT-4V'', ``Gemini'', ``Qwen-VL'', and ``CogVLM''. These terms were combined with ``multi-turn'', ``conversational'', ``context'', or ``history''.

\noindent\textbf{Training and alignment.}
Training-related keywords included ``instruction tuning'', ``instruction following'', ``reinforcement learning from human feedback'', ``RLHF'', ``direct preference optimisation'', ``DPO'', ``LoRA'', ``low-rank adaptation'', ``QLoRA'', ``parameter-efficient fine-tuning'', ``PEFT'', ``alignment'', ``constitutional AI'', ``RLAIF'', ``context window'', ``long context'', ``retrieval augmented generation'', ``RAG'', and ``memory-augmented''. We combined these with ``dialogue'', ``conversational'', ``multi-turn'', ``chatbot'', or ``assistant''.

\noindent\textbf{Evaluation frameworks and metrics.}
Evaluation-related searches used ``dialogue evaluation'', ``conversation evaluation'', ``turn-level evaluation'', ``multi-turn evaluation'', ``coherence'', ``engagement'', ``consistency'', ``factual grounding'', ``hallucination'', ``faithfulness'', ``dialogue benchmark'', ``conversational benchmark'', ``human evaluation dialogue'', ``automatic evaluation dialogue'', ``BLEU dialogue'', ``BERTScore dialogue'', ``perplexity dialogue'', ``FED'', ``USR'', ``GRADE'', ``DialEval'', and ``CGPR''.

\noindent\textbf{Cultural and multilingual grounding.}
We used ``multilingual dialogue'', ``cross-lingual dialogue'', ``Arabic dialogue'', ``Arabic conversational'', ``low-resource dialogue'', ``culturally grounded'', ``cultural dialogue'', ``dialect'', ``code-switching'', ``multilingual QA'', ``cross-cultural conversation'', ``Arabic NLP'', ``Arabic language model'', ``Arabizi'', ``cultural benchmark'', ``cultural bias'', ``cultural adaptation'', ``Gulf dialect'', ``MSA dialogue'', and ``Arabic ASR''. These terms were combined with ``multi-turn'', ``dialogue'', ``conversational'', ``chatbot'', or ``assistant''.

\section{Comparison with Related Surveys}

We extend the comparison in Table~\ref{tab:survey-positioning} in Table~\ref{tab:survey-positioning-detailed}, where we examine the conceptual framing, modality coverage, and key distinctions of closely related surveys. We show how our survey centers the complete multi-turn session and analyzes session-level competence across modalities, training, evaluation, and cultural--linguistic settings.

%% file: tables/multi-lingual-cultural_table.tex
\begin{table*}[!ht]
\centering
\small
\setlength{\tabcolsep}{2.5pt}

\label{tab:cultural-resources}
\resizebox{\textwidth}{!}{%
\begin{tabular}{l l l l c c c l}
\toprule
\textbf{Resource} & \textbf{Year} & \textbf{Languages} & \textbf{Region / Cultures} & \textbf{MT} & \textbf{MM} & \textbf{Dialect} & \textbf{Venue} \\
\midrule
\multicolumn{8}{c}{\textit{\textbf{Arabic / MENA}}} \\
\midrule

\sys{OASIS / EverydayMMQA}~\cite{alam2025everydaymmqa}
  & 2025 & EN, MSA, Egyptian, Levantine & 18 Arab countries & \xmark$^{\star}$ & \cmark & \cmark & arXiv \\
\sys{M$^2$CQA}~\cite{mousi2026m2cqa}
  & 2025 & MSA + dialects               & 17 MENA countries & \xmark & \cmark & \cmark & arXiv \\

\sys{Dallah}~\cite{alwajih2024dallah}
  & 2024 & 6 Arabic dialects            & MENA              & \xmark & \cmark & \cmark & ArabicNLP \\
\sys{Arabic Dialect Dialogue}~\cite{naous2020empathy}
  & 2020 & Levantine, Egyptian, Gulf    & MENA              & \cmark & \xmark & \cmark & WANLP \\

\sys{ArabicaQA}~\cite{abdallah2024arabicaqa}
  & 2024 & Arabic (MSA)                 & MENA              & \xmark & \xmark & \xmark & SIGIR \\
\sys{MENA Speech Bank}~\cite{ali2026menaspeechbankreferencevoicebank}
  & 2026 & Arabic + dialects            & MENA              & \cmark & \cmark & \cmark & arXiv \\
\sys{Shawarma Chats}~\cite{zeinalipour-etal-2025-shawarma} & 2025 & MSA, Egyptian, Maghrebi & MENA & \cmark & \xmark & \cmark & ArabicNLP \\
\sys{Alexandria}~\cite{MMAA2026} & 2026 & en + 13 Arabic dialects & 13 Arab countries & \cmark & \xmark & \cmark & ACL \\
\midrule
\multicolumn{8}{c}{\textit{\textbf{Chinese}}} \\
\midrule

\sys{AlignMMBench}~\cite{wu2025alignmmbench}
  & 2025 & zh                           & China             & \cmark & \cmark & \xmark & ACL \\
\sys{CMT-Eval}~\cite{tian2025cmt}
  & 2025 & zh                           & China             & \cmark & \xmark & \cmark & arXiv \\

\sys{FB-Bench}~\cite{li2025fb}
  & 2025 & zh + en                      & China             & \cmark & \xmark & \xmark & EMNLP \\
\midrule
\multicolumn{8}{c}{\textit{\textbf{Multilingual}}} \\
\midrule

\sys{CulturalGround}~\cite{nyandwi2025grounding}
  & 2025 & 39 langs                     & 42 countries      & \xmark & \cmark & --     & EMNLP \\
\sys{CVQA}~\cite{mogrovejo2024cvqa}
  & 2024 & 26 langs                     & 30 countries      & \xmark & \cmark & --     & NeurIPS \\

\sys{MMA-ASIA}~\cite{weihua2025mmaasia}
  & 2025 & 10 langs (8 Asian countries) & Asia              & \xmark & \cmark & --     & arXiv \\
\sys{mSTEB}~\cite{beyene2025msteb}
  & 2025 & 40+ langs                    & --                & \xmark & \cmark & --     & arXiv \\

\sys{M2Lingual}~\cite{maheshwary2025m2lingual}
  & 2025 & ML                           & --                & \cmark & \xmark & --     & arXiv \\
\sys{WildChat}~\cite{zhao2024wildchat}
  & 2024 & ML (incl.\ code-switch)      & 200+ countries    & \cmark & \xmark & --     & ICLR \\
\sys{MLQA}~\cite{lewis-etal-2020-mlqa}
  & 2020 & en, ar, de, es, hi, vi, zh & Cross-lingual & \xmark & \xmark & -- & ACL \\
\sys{ADU-Bench}~\cite{gao-etal-2025-benchmarking} & 2025 & ar, zh, en, fr, de, ja, ko, ru, es & Multilingual audio dialogue & \xmark & \cmark & \xmark & ACL \\
\sys{MT-Bench-Hi}~\cite{kamath-etal-2025-benchmarking} & 2025 & Hindi & India & \cmark & \xmark & \xmark & ACL-W \\
\sys{cuDialog}~\cite{cao-etal-2024-bridging} & 2024 & en & Multiple cultures & \cmark & \xmark & \xmark & EACL-F \\
\sys{IndoToD}~\cite{kautsar-etal-2023-indotod} & 2023 & Indonesian & Southeast Asia & \cmark & \xmark & \xmark & SEALP \\
\sys{Typhoon-Audio}~\cite{manakul2025enhancing} & 2025 & Thai, English & Thailand / Thai speech & \xmark & \cmark & \xmark & Interspeech \\
\sys{SEA-VQA}~\cite{urailertprasert-etal-2024-sea} & 2024 & en & Southeast Asia / 8 countries, 53 
cultures & \xmark & \cmark & \xmark & ALVR \\
\bottomrule
\end{tabular}
}
\vspace{-0.3cm}
\caption{%
  \textbf{Cultural and linguistic resources relevant to multi-turn / multimodal dialogue.}
  \textbf{MT}: \cmark\,=\,multi-turn structure; \xmark\,=\,single-turn only.
  \textbf{MM}: \cmark\,=\,more than one modality (vision and/or speech).
  \textbf{Dialect}: \cmark\,=\,dialect-aware (beyond MSA or a single  variety).  
}
\vspace{-0.4cm}
\end{table*}

%% file: tables/modelling-architecture_table.tex
\begin{table*}[!ht]
\centering
\small
\setlength{\tabcolsep}{3.5pt}
\resizebox{\textwidth}{!}{%
\begin{tabular}{l l l l c l l}
\toprule
\textbf{Model} & \textbf{Year} & \textbf{Mod.} & \textbf{Ctx} & \textbf{Stream} & \textbf{Key novelty} & \textbf{Venue} \\
\midrule
\multicolumn{7}{c}{\textit{\textbf{Pre-LLM transformer dialogue managers}}} \\
\midrule

\sys{DialoGPT}~\cite{zhang-etal-2020-dialogpt}      & 2020 & T   & CC  & \xmark & 147M Reddit pre-train                   & ACL \\
\sys{PLATO}~\cite{bao-etal-2020-plato}              & 2020 & T   & CC  & \xmark & Discrete latent for diversity           & ACL \\

\sys{ToD-BERT}~\cite{wu2020tod}          & 2020 & T   & CC  & \xmark & Unified TOD pre-train (9 corpora)       & EMNLP \\
\sys{BlenderBot}~\cite{roller-etal-2021-recipes} & 2021 & T   & CC  & \xmark & Blended persona / knowledge / empathy   & EACL \\

\sys{TurnGPT}~\cite{ekstedt-skantze-2020-turngpt}      & 2020 & T+S & CC  & \xmark & Turn-taking transformer                 & SIGDIAL \\
\sys{PPTOD}~\cite{su2022multi}               & 2022 & T   & CC  & \xmark & Plug-and-play multi-task TOD            & ACL \\

\sys{KA-GPT2}~\cite{lin-etal-2021-knowledge}
  & 2021 & T & CC & \xmark & KG-aware GPT-2 for dialogue state tracking & EMNLP \\

\sys{Commonsense+NER}~\cite{varshney2022commonsense}
  & 2022 & T & CC & \xmark & KG + NER grounding for open-domain dialogue & NAACL \\
\sys{DLGNet}~\cite{oluwatobi-mueller-2020-dlgnet}
  & 2020 & T & CC & \xmark & Contextual Transformer encoding for response generation & ACL-W \\
\sys{MT-BERT-DST}~\cite{kapelonis22_interspeech}
  & 2022 & T & CC & \xmark & Shared BERT encoder; multi-task intent + slot prediction & Interspeech \\
  
\midrule
\multicolumn{7}{c}{\textit{\textbf{LLM-based, text-only}}} \\
\midrule

\sys{InstructGPT}~\cite{ouyang2022training}  & 2022 & T   & CC  & \xmark & RLHF (PPO) alignment                   & NeurIPS \\
\sys{COMEDY}~\cite{chen2025compress}         & 2025 & T   & MA  & \xmark & Compressive session memory              & COLING \\
\sys{FnCTOD}~\cite{li2024large} & 2024 & T & CC & \xmark & Zero-shot DST via function calling; no task-specific training & ACL \\
\sys{UniConv}~\cite{mo-etal-2025-uniconv} & 2025 & T & CC & \xmark & Joint retrieval-generation model for multi-turn conversational search & ACL \\
\sys{ChatQA}~\cite{liu2024chatqa} & 2024 & T & CC & \xmark & LLM + dedicated dense retriever jointly tuned for multi-turn conversational QA & NeurIPS \\

\midrule
\multicolumn{7}{c}{\textit{\textbf{Long-context / recurrent architectures}}} \\
\midrule

\sys{Transformer-XL}~\cite{dai2019transformerxl} & 2019 & T & RNN & \xmark & Segment-level recurrence              & ACL \\
\sys{RMT}~\cite{bulatov2022recurrent}              & 2022 & T   & MA  & \xmark & Memory tokens                          & NeurIPS \\

\sys{RWKV}~\cite{peng2023rwkv}               & 2023 & T   & RNN & \xmark & Linear-attn RNN-like                   & arXiv \\
\sys{MemBART}~\cite{membart2024}        & 2024 & T   & MA  & \xmark & Dual-attn read/write per turn           & arXiv \\
\sys{LLM-DST}~\cite{feng-etal-2023-towards}
  & 2023 & T & CC & \xmark & LLM-driven dialogue state tracking; & EMNLP \\
\sys{CCM}~\cite{kim2024compressed}
  & 2024 & T & MA & \xmark & KV-cache compression into compact context memory for online LM interaction & ICLR \\
  
\midrule
\multicolumn{7}{c}{\textit{\textbf{AudioLLMs (speech + text)}}} \\
\midrule

\sys{SpeechGPT}~\cite{zhang2023speechgpt}    & 2023 & T+S & CC & \xmark & Discrete speech tokens                 & EMNLP-F \\
\sys{SALMONN}~\cite{tang2024salmonn}         & 2024 & T+S & CC & \xmark & Dual audio encoder                     & ICLR \\

\sys{Qwen2-Audio}~\cite{chu2024qwen2}   & 2024 & T+S & CC & \xmark & Multi-task audio chat                  & arXiv \\
\sys{Moshi}~\cite{defossez2024moshi}    & 2024 & T+S & CC & \cmark & Full-duplex; 160ms; inner monologue    & arXiv \\
\sys{AudioPaLM}~\cite{rubenstein2023audiopalm}
  & 2023 & T+S & CC & \xmark & PaLM extended with audio tokens for speech + text & arXiv \\
\sys{Qwen-Audio}~\cite{chu2023qwenaudio}
  & 2023 & T+S & CC & \xmark & Universal audio understanding via unified audio-language model & arXiv \\
\sys{Style-Talker}~\cite{li2024styletalkerfinetuningaudiolanguage}
  & 2024 & T+S & CC & \xmark & Audio-LLM + style-based TTS for fast spoken dialogue & arXiv \\
\sys{Mini-Omni}~\cite{xiezhifei2024miniomni}
& 2024 & T+S & CC & \cmark & End-to-end real-time speech interaction with parallel text-audio streaming output & arXiv \\
\sys{Freeze-Omni}~\cite{wang2025freezeomni}
  & 2025 & T+S & CC & \cmark & Low-latency speech dialogue with frozen LLM & ICML \\
\sys{MinMo}~\cite{chen2025minmomultimodallargelanguage}
  & 2025 & T+S & CC & \cmark & Full-duplex multimodal LLM for seamless spoken interaction & arXiv \\
\sys{dGSLM}~\cite{nguyen-etal-2023-generative} & 2023 & S & CA & \xmark & First textless spoken dialogue model; dual-tower cross-attention on raw two-channel audio & TACL \\
\sys{LLaMA-Omni 2}~\cite{fang2025llamaomni2} & 2025 & T+S & CC & \cmark & Autoregressive streaming speech synthesis for real-time spoken chatbot & ACL \\
\sys{SpiRit-LM}~\cite{nguyen-etal-2025-spirit} & 2025 & T+S & CC & \xmark & Interleaved speech-text tokens; word-level alignment; prosody preserved & TACL \\
\sys{GLM-4-Voice}~\cite{zeng2024glm4voice} & 2024 & T+S & CC & \cmark & End-to-end spoken chatbot; controllable emotion, rate, and dialect & arXiv \\
\sys{SLAM-Omni}~\cite{chen2025slam} & 2025 & T+S & KV & \cmark & Historical text prompting; timbre control; single-stage SDM training & ACL-F \\
\sys{Step-Audio~2}~\cite{wu2025stepaudio2} & 2025 & T+S & CC & \cmark & Large-scale audio-LM; streaming speech, RAG, and tool use & arXiv \\

\midrule
\multicolumn{7}{c}{\textit{\textbf{Omni-modal (text + speech + vision)}}} \\
\midrule

\sys{GPT-4o}~\cite{hurst2024gpt}           & 2024 & T+S+V & CC & \cmark & End-to-end omni baseline              & OpenAI \\
\sys{Qwen2.5-Omni}~\cite{qwen25omni}  & 2025 & T+S+V & CC & \cmark & Thinker--Talker + TMRoPE             & arXiv \\

\sys{VITA-1.5}~\cite{fu2026vita}          & 2026 & T+S+V & CC & \cmark & 1.5s end-to-end latency              & NeurIPS \\
\sys{IXC2.5-OmniLive}~\cite{zhang2024ixcomnilive} & 2024 & T+S+V & MA & \cmark & Streaming + long-term memory      & arXiv \\

\sys{Ming-Omni}~\cite{ai2025ming}     & 2025 & T+S+V & CC & \cmark & Unified perception + generation      & arXiv \\
\sys{InteractiveOmni}~\cite{tong2025interactiveomni}
  & 2025 & T+S+V & MA & \cmark & Tri-modal audio-visual multi-turn dialogue; cross-modal memory & arXiv \\

\sys{Mini-Omni2}~\cite{xie2024miniomni2opensourcegpt4ovision}
  & 2024 & T+S+V & CC & \cmark & Adds vision to Mini-Omni; full duplex & arXiv \\

\sys{EMOVA}~\cite{chen2025emova}
  & 2025 & T+S+V & CC & \cmark & Disentangled speech tokeniser + vision encoder; vivid emotions & CVPR \\

\sys{Stream-Omni}~\cite{zhang2025stream}
  & 2025 & T+S+V & CC & \cmark & Simultaneous multimodal interactions; streaming & arXiv \\

\sys{M2-Omni}~\cite{guo2025m2}
  & 2025 & T+S+V & CC & \cmark & Comprehensive modality support with competitive performance & arXiv \\

\sys{Baichuan-Omni}~\cite{li2024baichuanomnitechnicalreport}
  & 2024 & T+S+V & CC & \cmark & Omni-modal technical report & arXiv \\

\sys{MIO}~\cite{wang-etal-2025-mio}
  & 2025 & T+S+V & CC & \cmark & Foundation model on multimodal tokens & EMNLP \\

\sys{Vision-Speech}~\cite{royer2025visionspeechmodels} & 2025 & T+S+V & CC & \xmark & Spoken dialogue grounded in vision; no text intermediary & arXiv \\
\sys{MGM-Omni}~\cite{wang2025mgm} & 2025 & T+S+V & CC & \cmark & Dual-track omni LLM; long-audio understanding and personalized long speech & arXiv \\
\sys{Ola}~\cite{liu2025ola} & 2025 & T+S+V & CC & \xmark & Omni-modal input understanding with progressive modality alignment & arXiv \\
\midrule
\multicolumn{7}{c}{\textit{\textbf{Multimodal multi-turn-aware (vision + text, context-managed)}}} \\
\midrule

\sys{ContextQFormer}~\cite{lei2025contextqformer}
  & 2025 & T+V & QC  & \xmark & Memory queue + cross-attention         & arXiv \\
\sys{MadaKV}~\cite{li-etal-2025-madakv}
  & 2025 & T+V & KV  & \xmark & Modality-aware KV-cache eviction       & ACL \\

\sys{DiagNote}~\cite{liu2025diagnote}
  & 2025 & T+V & MA  & \xmark & Note-taking module for VLMs            & EMNLP \\
\sys{Dallah}~\cite{alwajih2024dallah}
  & 2024 & T+V & CC  & \xmark & Dialect-aware Arabic VLM               & ACL \\
\sys{VideoLLaMA\,2}~\cite{damonlpsg2024videollama2}
  & 2024 & T+S+V & CC & \xmark & Audio branch extension of LLaMA for video+audio & arXiv \\

\sys{CAT}~\cite{ye2024cat}
  & 2024 & T+S+V & CC & \xmark & Dynamic audio-visual scene QA alignment under temporal change & ECCV \\

\sys{LoopServe}~\cite{li2025loopserve}
  & 2025 & T+V & KV & \xmark & Adaptive sparsification for multi-turn KV pressure at inference & arXiv \\

\sys{AirCache}~\cite{huang2025aircache}
  & 2025 & T+V & KV & \xmark & Inter-modal relevancy KV compression; elite observation window & ICCV \\
\sys{ClarifyWN}~\cite{zhang2025clarify}
  & 2025 & T & CA & \xmark & Clarification-as-policy; when/what to ask across turns & NAACL \\
\sys{DialogGen}~\cite{huang-etal-2025-dialoggen} 
& 2025 & T+V & CC & \xmark & Bilingual text-image generation and editing conditioned on multi-turn conversational history & NAACL-F \\

\midrule
\multicolumn{7}{c}{\textit{\textbf{Agentic / tool-augmented (adjacent scope)}}} \\
\midrule

\sys{ReAct}~\cite{yao2022react}              & 2023 & T   & CA  & -- & Reasoning + acting interleaved          & ICLR \\
\sys{Reflexion}~\cite{shinn2023reflexion}    & 2023 & T   & MA  & -- & Verbal RL via self-reflection           & NeurIPS \\

\sys{MetaGPT}~\cite{hong2024metagpt}         & 2024 & T   & CA  & -- & Role-specialised multi-agent            & ICLR \\
\sys{SAPIENT}~\cite{du2025sapient}           & 2025 & T   & CA  & -- & MCTS planner for conversational rec.   & NAACL \\
\sys{Toolplanner}~\cite{wu2024toolplanner}  & 2024 & T & CA & -- & Path planning + feedback RL  & EMNLP\\
\sys{CHATCOT}~\cite{chen2023chatcot} & 2023 & T & CA & -- & Multi-turn tool-augmented CoT reasoning & EMNLP\\
\sys{WebLINX}~\cite{pmlr-v235-lu24e}
  & 2024 & T & CC & -- & Real-website navigation; 2,300 demos, 150+ sites & ICML \\
\sys{VideoMind}~\cite{liu2026videomind}
  & 2026 & T+Vid & CA & -- & Planner→grounder→verifier→answerer with Chain-of-LoRA & ICLR \\
\sys{Video-MTR}~\cite{xie2025videomtrreinforcedmultiturnreasoning}
  & 2025 & T+Vid & CA & -- & Iterative segment selection; gated bi-level reward & arXiv \\
\bottomrule
\end{tabular}
}
\vspace{-0.3cm}
\caption{%
  \textbf{Representative multi-turn dialogue models by paradigm} (full taxonomy).
  \textbf{Ctx}: CC\,=\,full-context concat; SW\,=\,sliding window; CA\,=\,cross-attention;
  MA\,=\,memory-augmented; QC\,=\,Q-Former / token compression; KV\,=\,KV-cache mgmt.;
  RNN\,=\,recurrent / SSM.
  \textbf{Streaming}: \cmark\,=\,real-time / full-duplex capable.%
}
\label{tab:models-full}
\vspace{-0.3cm}
\end{table*}

%% file: tables/training_strategies_tables.tex
\begin{table*}[!ht]
\centering
\small
\setlength{\tabcolsep}{2.5pt}
\resizebox{\textwidth}{!}{%
\begin{tabular}{l l l c l l}
\toprule
\textbf{Method} & \textbf{Year} & \textbf{Type} & \textbf{MT-aware} & \textbf{Multi-turn novelty} & \textbf{Venue} \\
\midrule
\multicolumn{6}{c}{\textbf{(i) Supervised fine-tuning}} \\
\midrule
\sys{WildChat}~\cite{zhao2024wildchat}
  & 2024 & SFT & \cmark & 1M real opt-in convs; multilingual & ICLR \\
\sys{Parrot}~\cite{sun2024parrot}
  & 2024 & SFT+PO & \cmark & Parrot-Ask elicits anaphora/ellipsis in follow-ups; SFT + context-aware preference opt. & ACL \\
\sys{Aquila-Med}~\cite{zhao2024aqulia}
  & 2024 & SFT+RLHF+DPO & \cmark & Full-process medical dialogue recipe  & arXiv \\
\sys{Qilin-Med}~\cite{ye2023qilin}
  & 2024 & SFT+DPO & \cmark & Multi-stage knowledge injection for medical multi-turn dialogue & arXiv \\
\sys{Zhongjing}~\cite{yang2024zhongjing}
  & 2024 & SFT+RLHF & \cmark & Expert-feedback RLHF + real-world multi-turn medical dialogue & AAAI \\
\midrule
\multicolumn{6}{c}{\textbf{(ii) Multi-task learning}} \\
\midrule
\sys{DAMSEL}~\cite{chen-etal-2025-data}
  & 2025 & Multi-task learning & \cmark & Auxiliary task design for spoken QA with limited speech data & ACL-F \\
\midrule
\multicolumn{6}{c}{\textbf{(iii) Reinforcement learning \& preference optimisation}} \\
\midrule
\sys{InstructGPT}~\cite{ouyang2022training}
  & 2022 & PPO & \xmark & RLHF preference recipe & NeurIPS \\
\sys{DMPO}~\cite{shi2024direct}
  & 2024 & DPO (MT) & \cmark & Direct multi-turn preference optimisation & EMNLP \\
\sys{M-DPO / M-KTO}~\cite{ICLR2025_40eff167}
  & 2025 & DPO/KTO & \cmark & Trajectory-level reward assignment (multi-turn math agents) & ICLR \\
\sys{ITPO}~\cite{wang2026implicit}
  & 2026 & RL & \cmark & Implicit turn-wise policy optimisation & arXiv \\
\sys{SDPO}~\cite{kong2025sdpo}
  & 2025 & DPO (segment) & \cmark & Segment-level preference optimisation for social dialogue & ACL \\
\sys{DiaTool-DPO}~\cite{jung-etal-2025-diatool}
  & 2025 & DPO & \cmark & Multi-turn dialogue-control optimisation for tool use & SIGDIAL \\
\sys{ArCHer}~\cite{pmlr-v235-zhou24t}
  & 2024 & HRL & \cmark & Two-level (on and off-policy RL)  & ICML \\
\sys{REFUEL}~\cite{gao2025refuel}
  & 2025 & RLHF & \cmark & Efficient multi-turn RLHF; regresses relative-future rewards & ICLR \\
\sys{SWEET-RL}~\cite{sweetrl2025}
  & 2025 & RL & \cmark & Turn-wise advantage critic for collaborative reasoning & arXiv \\
\sys{ACT}~\cite{chen2025learning}
  & 2025 & DPO / self-training & \cmark & Action-level contrastive self-training (clarify vs.\ answer) & ICLR \\
\sys{SCoRe}~\cite{kumar2025training}
  & 2025 & RL & \cmark & Online RL for self-correction over multi-turn traces & ICLR \\
\sys{MT-RLHF}~\cite{shani2024multiturn}
  & 2024 & RL & \cmark & Mirror-descent policy opt.; Nash from conversation-level prefs & NeurIPS \\
\sys{JOSH}~\cite{lattimer-etal-2025-sparse}
  & 2025 & Self-alignment/RL & \cmark & Sparse-reward self-training in MultiWOZ-derived tool env & ACL-F \\
\sys{DoctorAgent-RL}~\cite{feng2026doctoragent}
  & 2026 & RL & \cmark & Multi-agent collaborative RL for multi-turn clinical dialogue & ICASSP \\
\sys{Offline RL Persona}~\cite{shea2023building}
  & 2023 & Offline RL & \cmark & Persona-break penalty, Persona consistency critic & EMNLP \\
\sys{Student-Sim}~\cite{scarlatos2025training}
  & 2025 & SFT+DPO & \cmark & Tutor training via distillation + DPO; simulated-student model scores preference pairs & AIED \\
\midrule
\multicolumn{6}{c}{\textbf{(iv) Synthetic data generation}} \\
\midrule
\sys{UltraChat}~\cite{ding2023ultrachat}
  & 2023 & Synthetic data & \cmark & 1.5M synthetic multi-turn convs via paired-LLM self-chat & EMNLP \\
\sys{MMDU-45k}~\cite{liu2024mmdu}
  & 2024 & Synthetic (+bench) & \cmark & GPT-4o-generated multi-turn multi-image; doubles as benchmark & NeurIPS \\
\sys{TMDialog}~\cite{lei2025contextqformer}
  & 2025 & Synthetic data & \cmark & GPT-4-generated multi-turn multimodal dialogue corpus & arXiv \\
\sys{MMDiag}~\cite{liu2025diagnote}
  & 2025 & Synthetic data & \cmark & GPT-assisted multi-turn multimodal dialogue with note annotations & EMNLP \\
\sys{Arabic SFT Corpus}~\cite{misbah2026fine}
  & 2026 & Synthetic data & \cmark & 43K Arabic synthetic multi-turn dialogues; blueprint for Arabic SFT & PLoS ONE \\
\sys{DeepDialogue}~\cite{deepdialogue2025}
  & 2025 & Synthetic data & \cmark & Synthetic multi-turn spoken dialogue with emotion-consistency control & arXiv \\
\sys{ToolDial}~\cite{shim2025tooldial}
  & 2025 & Synthetic data & \cmark & Multi-turn dialogue generation pipeline for tool-augmented LMs & ICLR \\
\sys{ConsistentChat}~\cite{chen2025consistentchat}
  & 2025 & Synthetic data & \cmark & Skeleton-guided generation with nine intent trajectories & EMNLP \\
\sys{DocTalk}~\cite{lee-etal-2025-doctalk}
  & 2025 & Synthetic (PT) & \cmark & Graph-based synthesis of 730k multi-turn dialogues from Wikipedia (pre-training) & SIGDIAL \\
\sys{Review-Instruct}~\cite{wu2025instruct}
  & 2025 & Synthetic data & \cmark & Ask-Respond-Review three-role pipeline for high-consistency data & ACL-F \\
\sys{Self-Talk}~\cite{ulmer-etal-2024-bootstrapping}
  & 2024 & Synthetic data & \cmark & LLM-simulated user-agent task dialogue generation & ACL-F \\
\sys{APIGen-MT}~\cite{prabhakar2026apigenmt}
  & 2026 & Synthetic data & \cmark & Verifiable multi-turn agent trajectory generation & NeurIPS \\
\midrule
\multicolumn{6}{c}{\textbf{(v) Conversational retrieval-augmented training}} \\
\midrule
\sys{ChatQA}~\cite{liu2024chatqa}
  & 2024 & SFT (retrieval) & \cmark & Two-stage tuning of retriever + LLM for multi-turn conversational QA & NeurIPS \\
\sys{ChatQA-2}~\cite{xu2025chatqa}
  & 2025 & SFT (retrieval) & \cmark & Three-stage context-expansion tuning (8K→128K) for long conversational RAG & ICLR \\
\sys{HAConvDR}~\cite{mo-etal-2024-history}
  & 2024 & Dense retrieval & \cmark & Context-denoised reformulation + turn-impact supervision & ACL-F \\
\sys{ConvAUG}~\cite{chen-etal-2024-generalizing}
  & 2024 & Dense retrieval & \cmark & LLM-cognition augmentation + difficulty-adaptive contrastive learning & ACL \\
\bottomrule
\end{tabular}}
\vspace{-0.3cm}
\caption{
  \textbf{Multi-turn training strategies}, grouped by the five families of \S5.
  \textbf{Type}: SFT\,=\,supervised fine-tuning; DPO/KTO\,=\,direct-preference / Kahneman--Tversky optimisation (MT\,=\,multi-turn, segment\,=\,segment-level); RLHF/PPO, RL, HRL\,=\,hierarchical RL; \emph{retrieval}\,=\,conversational-retrieval training; \emph{Synthetic}\,=\,data-generation pipeline;
  Combined types (e.g.\ SFT+DPO) mark hybrids that span families; each work is listed under its \emph{primary} contribution.
  \textbf{MT-aware}: \cmark if the method explicitly optimises across turns rather than treating each turn independently.
}
\label{tab:training-full}
\vspace{-0.3cm}
\end{table*}

%% file: tables/evaluation_table.tex
\begin{table*}[!ht]
\centering
\small
\setlength{\tabcolsep}{2pt}
\renewcommand{\arraystretch}{1}

\resizebox{\textwidth}{!}{%
\begin{tabular}{l l l c l l}
\toprule
\textbf{Metric / Framework} & \textbf{Type} & \textbf{What it measures} & \textbf{MT-spec.} & \textbf{Core limitation} & \textbf{Applied in} \\
\midrule

\multicolumn{6}{c}{\textbf{Surface-form and task-oriented}} \\
\midrule
BLEU~\cite{papineni2002bleu}
& NLG & n-gram precision & \xmark & No semantics; no coherence signal & MT-Bench, CoQA \\
ROUGE~\cite{lin2004rouge}
& NLG & n-gram recall & \xmark & Surface-form only; ignores context & SpokenWOZ \\
BERTScore~\cite{zhang2019bertscore}
& NLG & Semantic embedding similarity & \xmark & Single utterance; history-blind & MMDU \\
MM-Relevance~\cite{feng2023mmdialog}
& NLG & Modality-aware response relevance & \cmark & Defined for two-modal only & MMDialog \\
JGA
& DST & Joint goal accuracy across slots & \cmark & Brittle to surface paraphrase & MultiWOZ \\
Slot-F1
& DST & Per-slot precision and recall & \xmark & Slot-type bias; misses long-range context & SpokenWOZ \\
Inform / Success
& DST & Entity provision and task success & \cmark & Binary; misses partial success & MultiWOZ \\

\midrule
\multicolumn{6}{c}{\textbf{Session-level metrics}} \\
\midrule
APR \& ARS~\cite{deshpande-etal-2025-multichallenge,gosai2025audiomultichallenge}
& CONS & Average pass rate and rubric score & \cmark & Turn-level and session-level scores can diverge & MultiChallenge %
\\
PWC~\cite{li2025firm}
& CONS & Position-weighted consistency & \cmark & Requires adversarial probe set & MT-Eval style \\
MMRC 6-axis~\cite{xue2025mmrc}
& MEM & Extract, reason, update, manage, recall, refuse & \cmark & Image management is rare in text-only settings & MMRC \\
TURNWISE gap~\cite{graf2026turnwise}
& CONS & Matched single-turn and multi-turn gap & \cmark & Requires paired single-turn references & General LLMs \\
WCSR~\cite{li2025structflowbench}
& CONS & Structural and intra-turn constraint satisfaction & \cmark & Depends on constraint extraction and LLM judging & StructFlowBench \\
MT-Eval patterns~\cite{kwan2024mteval}
& CONS & Recollection, expansion, refinement, follow-up & \cmark & Single-session; no cross-session coverage & MT-Eval \\
IHEval~\cite{zhang2025iheval}
& CONS & Instruction hierarchy across system, user, history, and tool outputs & \cmark & Conflict resolution remains difficult & IHEval \\
Feedback-loop rule inference~\cite{zhang-etal-2025-turnbench}
& CONS & Integration of structured feedback across turns & \cmark & Game setting may not cover open-domain dialogue & TurnBench-MS \\
DiaHalu taxonomy~\cite{chen-etal-2024-diahalu}
& CONS & Dialogue-level hallucination subtypes & \cmark & Subtype labels require human annotation & DiaHalu \\
PIF~\cite{epstein2024mmmtif}
& CONS & Programmatic instruction following across accumulated constraints & \cmark & Checks format constraints, not answer correctness & MMMT-IF \\

\midrule
\multicolumn{6}{c}{\textbf{Agentic and retrieval-grounded}} \\
\midrule
CORAL citation labeling~\cite{cheng-etal-2025-coral}
& AGENT & Retrieval, generation, and citation attribution & \cmark & Citation annotation requires human labeling & CORAL \\
$\tau$-pass$^k$~\cite{yao2025tau}
& AGENT & Task pass rate across repeated trials & \cmark & Requires live user simulator; high cost & $\tau$-Bench \\
ToolSandbox scoring~\cite{lu-etal-2025-toolsandbox}
& AGENT & Stateful trajectory with milestone and minefield scoring & \cmark & Simulator-dependent; limited domains & ToolSandbox \\
AgentBoard progress rate~\cite{ma2024agentboard}
& AGENT & Subgoal completion across agent trajectories & \cmark & Requires task-specific progress decomposition & AgentBoard \\

\midrule
\multicolumn{6}{c}{\textbf{Speech-native and full-duplex}} \\
\midrule
ADU-Bench~\cite{gao-etal-2025-benchmarking}
& JUDGE & Open-ended audio dialogue skills and ambiguities & \cmark & Judge-dependent; rubric design central & ADU-Bench \\
MTalk-Bench protocol~\cite{du2025mtalk}
& JUDGE+HUMAN & Pairwise and rubric-based S2S scoring & \cmark & Small ranking gaps unstable & MTalk-Bench \\
ASK-QA outcome similarity~\cite{chen-etal-2025-data}
& CONS & Semantic similarity after spoken clarification turns & \cmark & Uses simulated users and TTS follow-ups & ASK-QA \\
URO-Bench~\cite{yan-etal-2025-uro}
& CONS & Correctness, speech quality, speech-text match, latency & \cmark & Dimensions are not always separable & URO-Bench \\
FD-Bench interruption set~\cite{peng25b_interspeech}
& FD & SIR, SRIR, EIR, NIR, and SRR & \cmark & Synthetic user speech; limited systems & FD-Bench \\
FD-Bench timing set~\cite{peng25b_interspeech}
& FD & IRD, FSED, ERT, and EIT & \cmark & Requires reliable VAD, ASR, and audio logs & FD-Bench \\
Full-Duplex-Bench-v3~\cite{lin2026fdb_v3}
& FD+AGENT & Tool-use correctness and disfluency tolerance & \cmark & Emerging full-duplex tool-use setting & Full-Duplex-Bench-v3 \\

\midrule
\multicolumn{6}{c}{\textbf{LLM-as-judge and human evaluation}} \\
\midrule
MT-Bench~\cite{zheng2023judging}
& JUDGE & Pairwise preference and 1--10 rating & \cmark & Self-enhancement and verbosity bias & MT-Bench \\
GPTScore~\cite{fu2024gptscore}
& JUDGE & Scoring through instruction-prompt likelihood & \cmark & Prompt-sensitive; weak dialogue grounding & Multi-turn dialogue quality \\
LLM-Eval Analysis~\cite{zhang2024comprehensive}
& JUDGE & Coherence, engagement, informativeness & \cmark & No multi-turn-specific axes & Open-domain dialogue \\
BotChat~\cite{duan2024botchat}
& JUDGE & Human-likeness discrimination & \cmark & Confounded by GPT-4 dominance & Open LLMs \\
JudgeLM~\cite{zhu2025judgelm}
& JUDGE & Fine-tuned open judge & \cmark & Knowledge and position bias & JudgeLM bench \\
ContextualJudgeBench~\cite{xu2025does}
& JUDGE & Judge consistency under context and RAG & \cmark & Contextual sensitivity; reference dependence & RAG + multi-turn eval \\
Multi-Judge Evaluator~\cite{tang2025learning}
& JUDGE & Distilled agreement from multiple judges & \cmark & Dependent on judge pool quality & Multi-turn dialogue \\
ABC-Eval~\cite{11463289}
& HUMAN & Fine-grained behavioural criteria & \cmark & Labour-intensive; rater drift & Open-domain chat \\
MMDU rubric judging~\cite{liu2024mmdu}
& HUMAN & Long-form multimodal rubric judgment & \cmark & Expensive; hard to scale & MMDU \\
WildBench~\cite{lin2025wildbench}
& JUDGE & Pairwise evaluation on real user conversations & \cmark & Scores individual responses pairwise & WildBench \\

\bottomrule
\end{tabular}}
\vspace{-0.2cm}
\caption{
\textbf{Representative evaluation metrics and frameworks} for multi-turn dialogue.
\textbf{Type}: NLG\,=\,surface-form overlap; DST\,=\,task-state tracking;
CONS\,=\,consistency; MEM\,=\,memory; AGENT\,=\,agentic/stateful;
JUDGE\,=\,LLM-as-judge; HUMAN\,=\,human rubric; FD\,=\,full-duplex speech.
\textbf{MT-spec.}: \cmark if the metric is undefined or degenerate in single-turn settings.
}
\label{tab:eval-metrics-full}
\vspace{2pt}
\begin{minipage}{0.98\textwidth}
\footnotesize
\textit{FD-Bench metric abbreviations:}
SRR\,=\,successful reply rate;
SIR\,=\,successful interrupt rate;
SRIR\,=\,successful reply-to-interrupt rate;
EIR\,=\,early interrupt rate;
NIR\,=\,noise interrupt rate;
IRD\,=\,interrupt response delay;
FSED\,=\,first speech emit delay;
ERT\,=\,early reply time;
EIT\,=\,early interrupt time.
\end{minipage}
\vspace{-0.3cm}
\end{table*}

\begin{table*}[t]
\centering
\footnotesize
\setlength{\tabcolsep}{4pt}

\scalebox{0.95}{
\begin{tabular}{@{}p{2.6cm} p{3.8cm} p{3.2cm} p{4.4cm}@{}}
\toprule
\textbf{Evaluation family} & \textbf{What it measures} & \textbf{Best suited for} & \textbf{Main limitation} \\
\midrule

Surface-form \& task-oriented
& Response overlap, semantic similarity, dialogue-state accuracy, and task success
& Task-oriented dialogue and fast automatic scoring
& Limited signal on session-level behavior and sensitivity to valid paraphrases \\

Session-level
& Instruction retention, memory, constraint satisfaction, feedback integration, and dialogue-level hallucination
& Memory, consistency, and long-horizon interaction
& Metric definitions vary across benchmarks, limiting direct comparison \\

Agentic \& retrieval-grounded
& Tool use, stateful execution, evidence retrieval, and source attribution
& Tool-augmented agents and conversational RAG
& Limited support for hidden state, side effects, and repeated-trial reliability \\

Speech-native \& full-duplex
& Semantic quality, speech quality, paralinguistic cues, interruption handling, latency, and timing
& Spoken and full-duplex dialogue systems
& Semantic correctness, speech quality, and interaction timing are often evaluated separately \\

LLM-as-judge \& human
& Rubric-based and pairwise judgments of open-ended response and dialogue quality
& Open-ended or subjective evaluation
& Sensitive to judge bias, rater inconsistency, cost, and reproducibility \\

\bottomrule
\end{tabular}
}
\vspace{-0.3cm}
\caption{Comparison of five evaluation families for multi-turn dialogue.}
\label{tab:eval-compare}
\vspace{-0.3cm}
\end{table*}